\def\year{2022}\relax
\documentclass[letterpaper]{article} 
\usepackage{aaai22}  
\usepackage{times}  
\usepackage{helvet}  
\usepackage{courier}  
\usepackage[hyphens]{url}  
\usepackage{graphicx} 
\usepackage{natbib}  
\usepackage{caption} 
\DeclareCaptionStyle{ruled}{labelfont=normalfont,labelsep=colon,strut=off} 
\usepackage{calc}
\usepackage{amsmath}
\usepackage{caption}
\usepackage{multirow}
\usepackage{amsfonts}
\usepackage[font=footnotesize]{subcaption}
\usepackage{booktabs}
\usepackage{pdfpages}
\usepackage{xcolor}
\usepackage{tabularx}
\usepackage{bbm}
\usepackage{graphicx}
\usepackage{algorithm}
\usepackage{algorithmic}
\newcommand{\answerYes}[1]{\textcolor{blue}{#1}} 
\newcommand{\answerNo}[1]{\textcolor{teal}{#1}} 
\newcommand{\answerNA}[1]{\textcolor{gray}{#1}}

\definecolor{weicolor}{rgb}{0.75,0.25,0}

\newcommand{\honeybee}{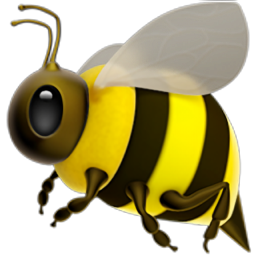}
\newcommand{\upsidedownface}{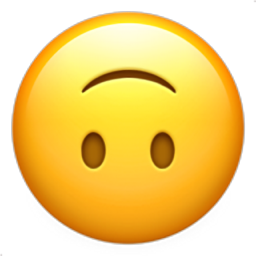}

\newcommand{\christmastree}{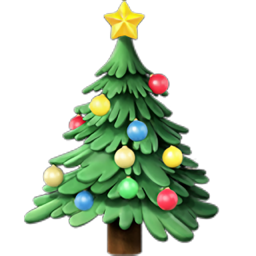}

\newcommand{\flexed}{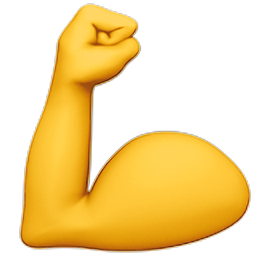}
\newcommand{\kiss}{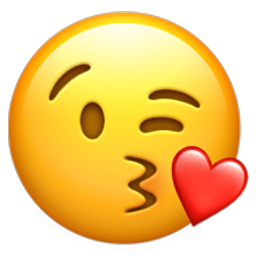}
\newcommand{\basketball}{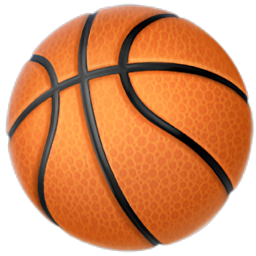}
\newcommand{\sparkles}{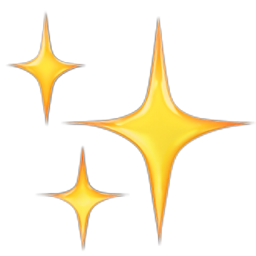}
\newcommand{\unamusedface}{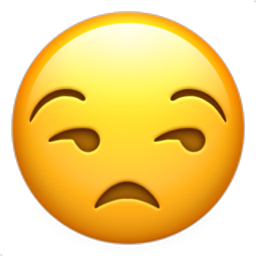}
\newcommand{\relieved}{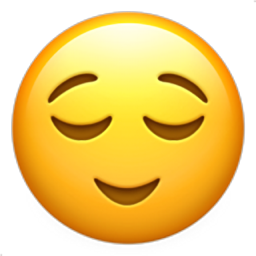}
\newcommand{\pleadingface}{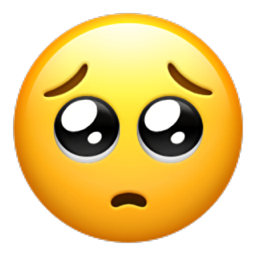}
\newcommand{\heart}{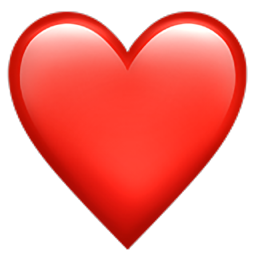}
\newcommand{\blackheart}{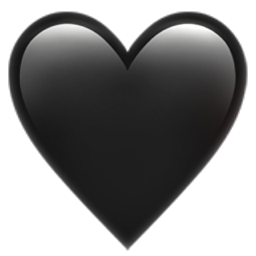}
\newcommand{\womandancing}{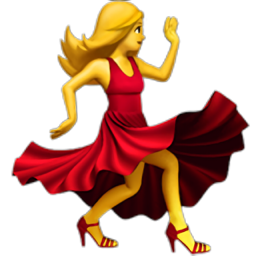}
\newcommand{\mandancing}{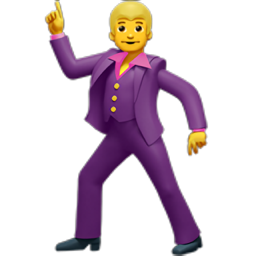}
\newcommand{\peoplewithbunnyears}{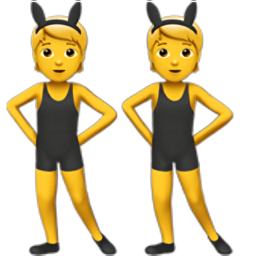}
\newcommand{\bouquet}{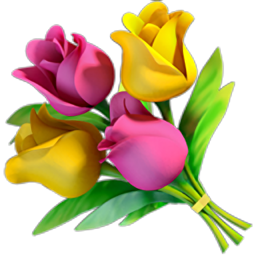}
\newcommand{\popcorn}{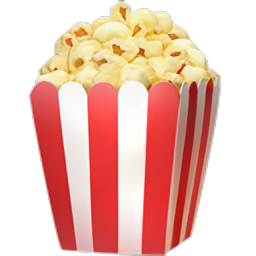}
\newcommand{\TV}{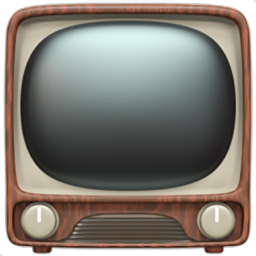}
\newcommand{\fire}{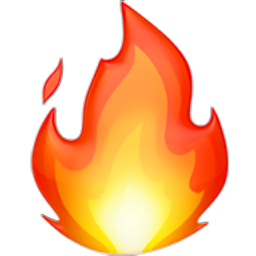}

\newcommand{\pray}{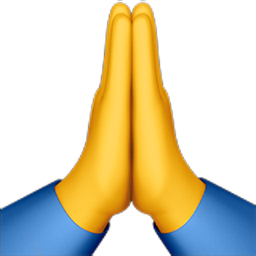}
\newcommand{\goat}{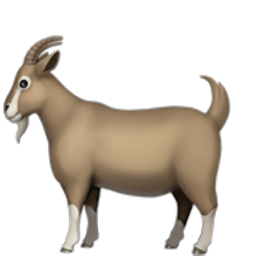}

\newcommand{\joy}{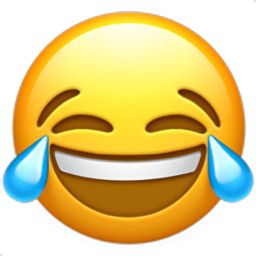}

\newcommand{\fistraised}{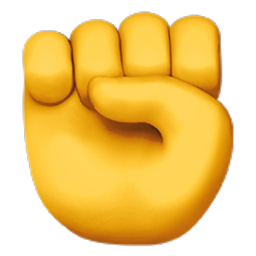}

\newcommand{\sob}{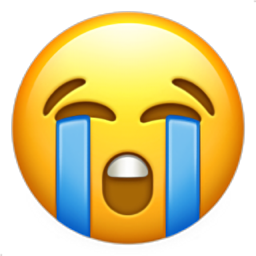}
\newcommand{\hearteyes}{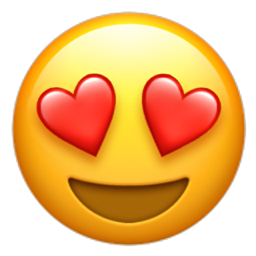}
\newcommand{\eyes}{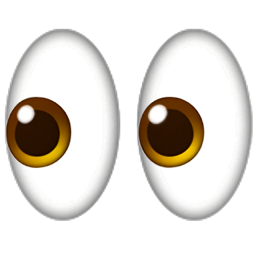}
\newcommand{\twohearts}{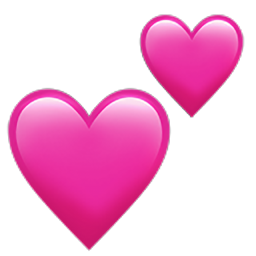}
\newcommand{\whiteheart}{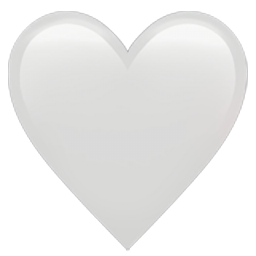}
\newcommand{\laptop}{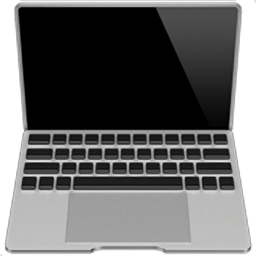}
\newcommand{\penguin}{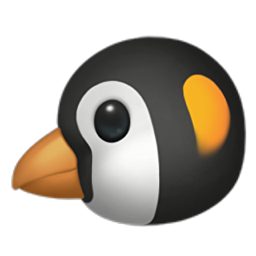}
\newcommand{\guitar}{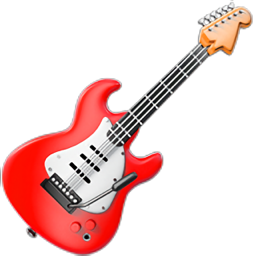}
\newcommand{\motorcycle}{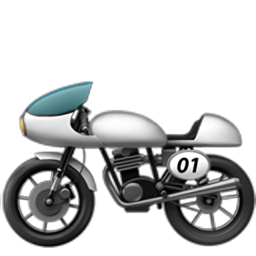}
\newcommand{\womanholdinghands}{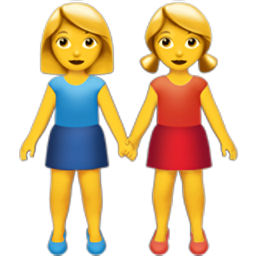}
\newcommand{\familywomangirlboy}{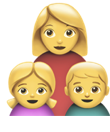}
\newcommand{\gift}{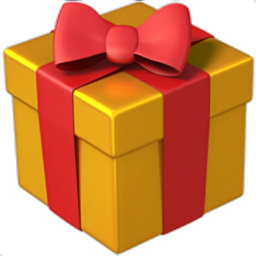}
\newcommand{\snake}{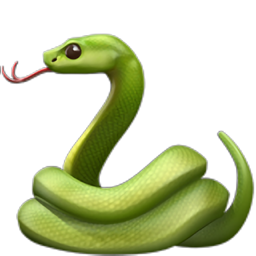}
\newcommand{\dizzy}{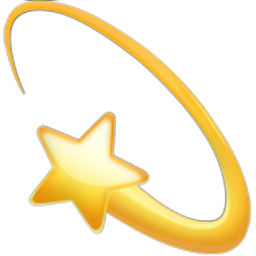}

\newcommand{\globalemoji}{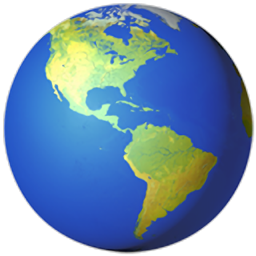}
\newcommand{\beardman}{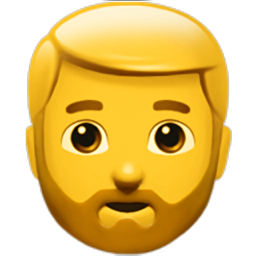}
\newcommand{\videogame}{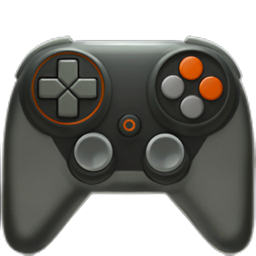}
\newcommand{\socceremoji}{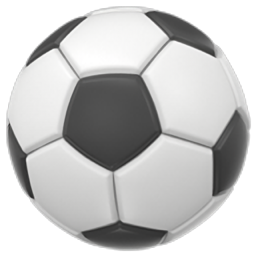}
\newcommand{\hamburger}{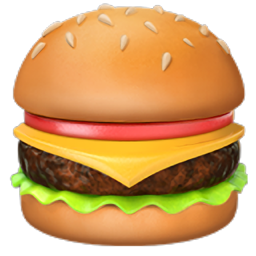}
\newcommand{\manlifting}{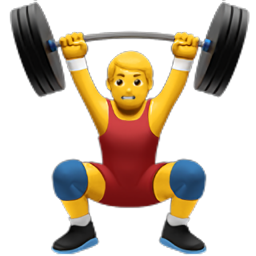}
\newcommand{\chart}{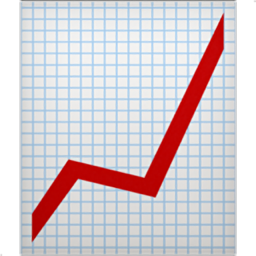}
\newcommand{\construction}{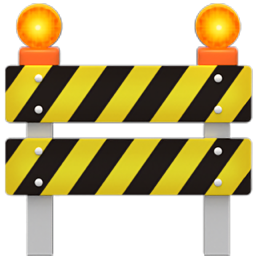}
\newcommand{\counterclockwisearrowsbutton}{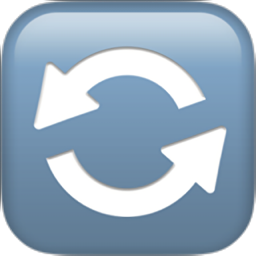}
\newcommand{\musicalnote}{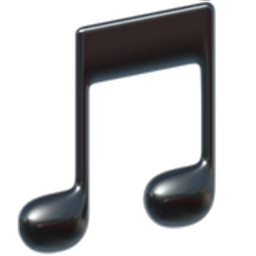}
\newcommand{\microphone}{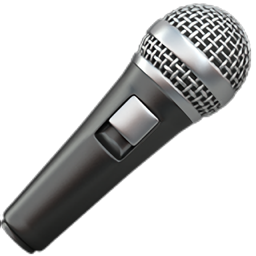}
\newcommand{\camerawithflash}{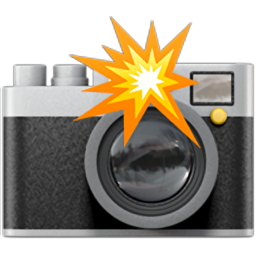}
\newcommand{\sparklingheart}{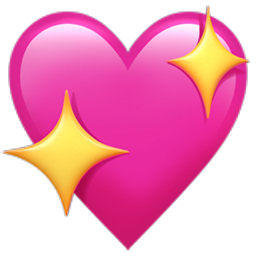}
\newcommand{\tower}{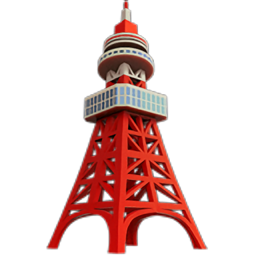}
\newcommand{\cheese}{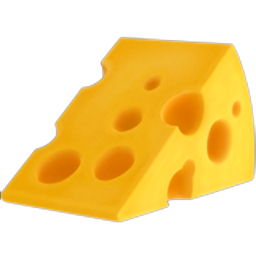}
\newcommand{\baguette}{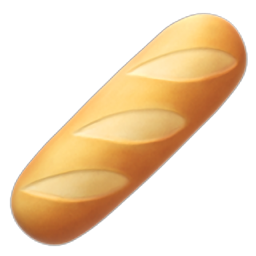}
\newcommand{\clinking}{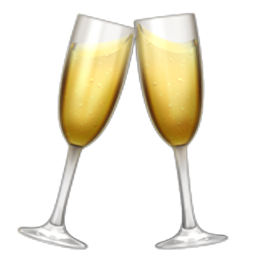}
\newcommand{\kissmark}{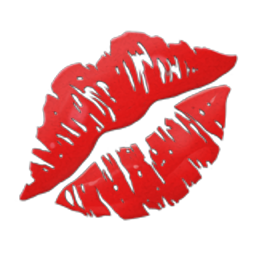}
\newcommand{\croissant}{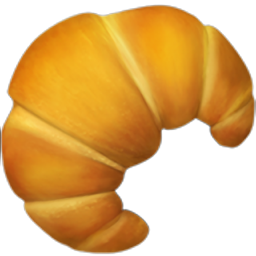}
\newcommand{\heartwitharrow}{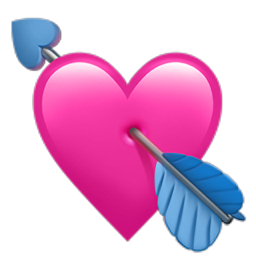}
\newcommand{\revolvinghearts}{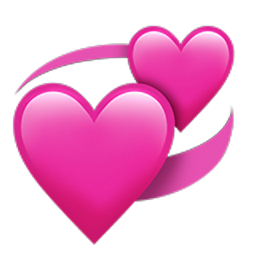}
\newcommand{\growingheart}{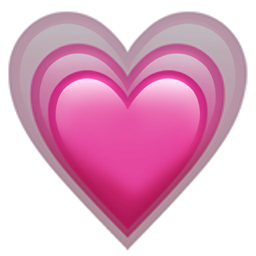}
\newcommand{\womanofficeworker}{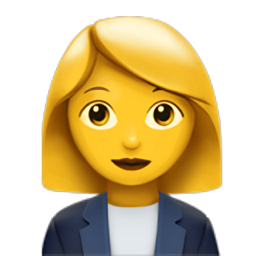}
\newcommand{\womanstudent}{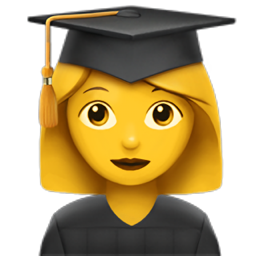}
\newcommand{\womansteamy}{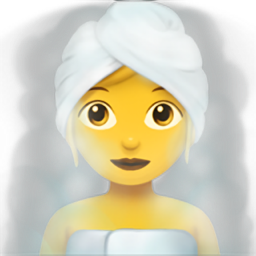}
\newcommand{\franceflag}{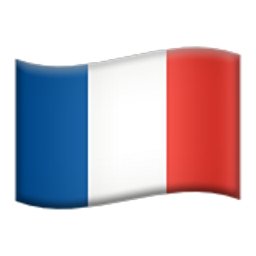}
\newcommand{\wineglass}{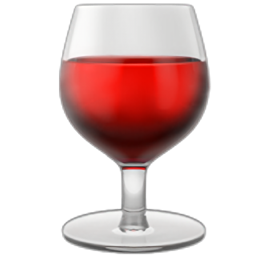}

\newcommand{\smilingfacewithsunglasses}{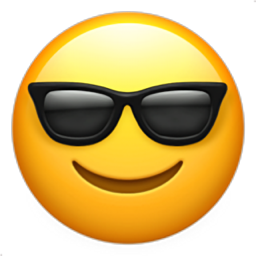}
\newcommand{\smirk}{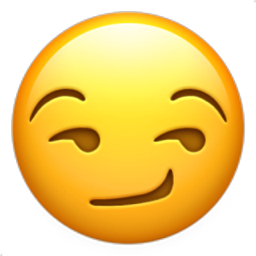}
\newcommand{\tippinghandwoman}{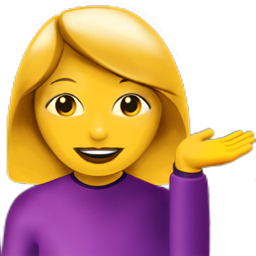}

\newcommand{\hammer}{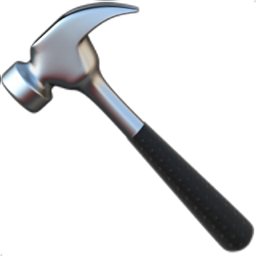}
\newcommand{\stuckouttonguewinkingface}{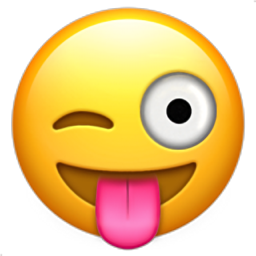}

\newcommand{\checkmarkemoji}{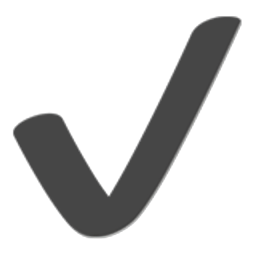}

\newcommand{\thumbsup}{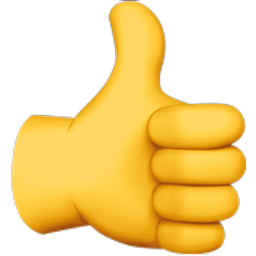}
\newcommand{\stopsign}{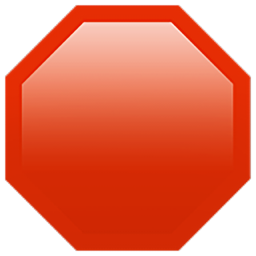}
\newcommand{\smilingeyes}{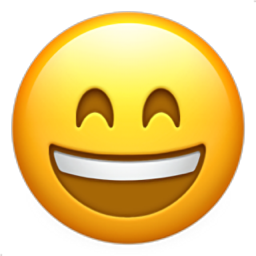}
\newcommand{\rocket}{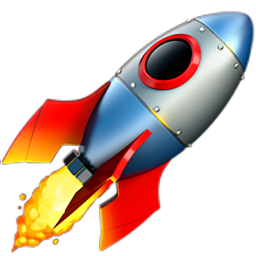}
\newcommand{\wavinghand}{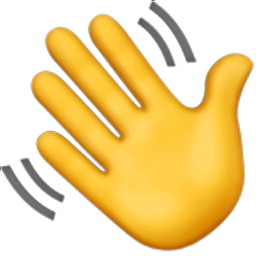}
\newcommand{\partypopper}{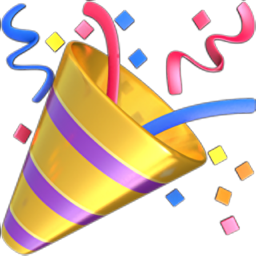}
\newcommand{\whitecheckmark}{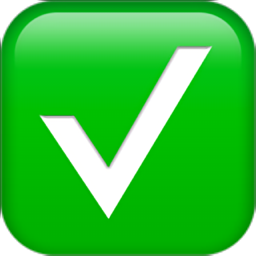}
\newcommand{\bugemoji}{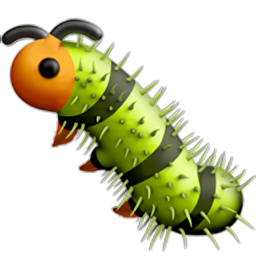}
\newcommand{\warningemoji}{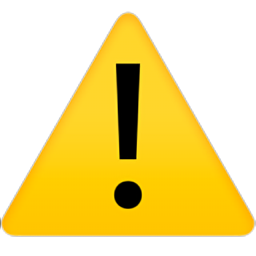}
\newcommand{\thinkingface}{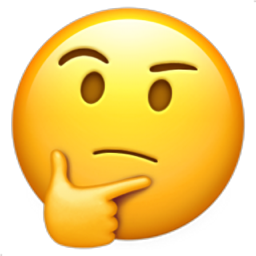}
\newcommand{\slightsmilingface}{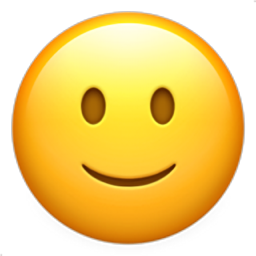}
\newcommand{\grinningface}{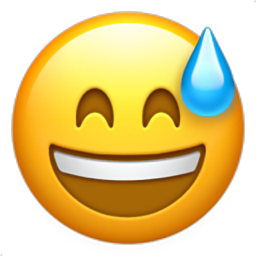}
\newcommand{\ladybugemoji}{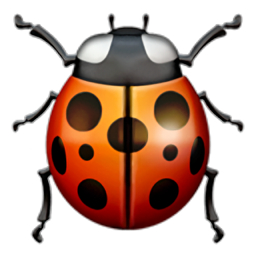}
\newcommand{\smilingfacesmilingeyes}{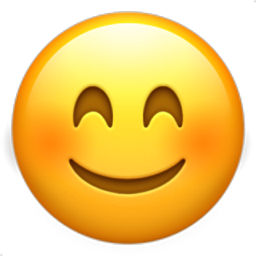}
\newcommand{\moneybag}{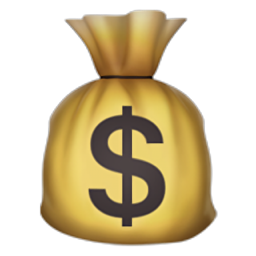}
\newcommand{\triangular}{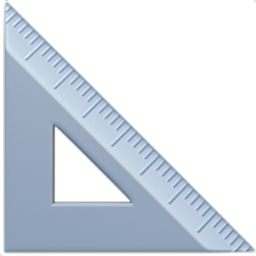}
\newcommand{\calendaremoji}{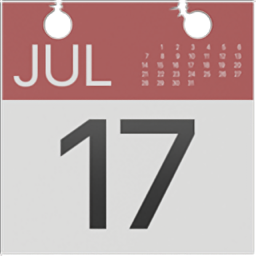}
\newcommand{\books}{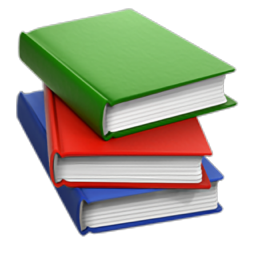}
\newcommand{\openbook}{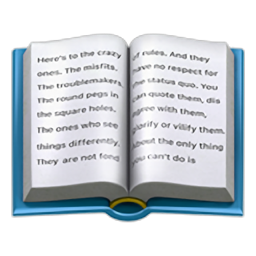}
\newcommand{\bigeyes}{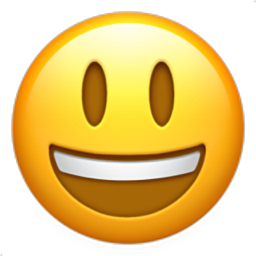}

\newcommand{\smilingfacewithhearts}{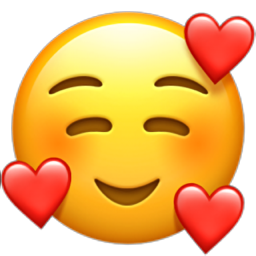}
\newcommand{\smilingfacewithsmilingeyes}{EmojiiFolder/u1F60A.png}

\newcommand{\shrugging}{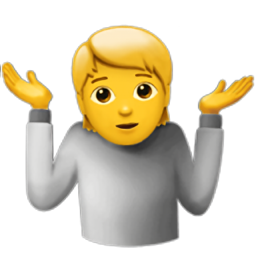}
\newcommand{\clipboard}{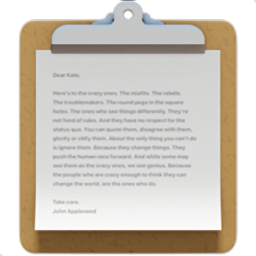}

\newcommand{\robot}{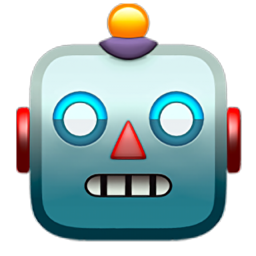}

\newcommand{\thumbsdown}{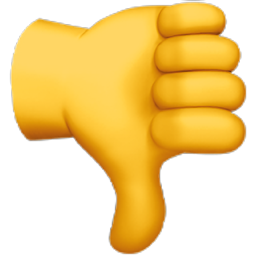}

\newcommand{\smilingfacewithtear}{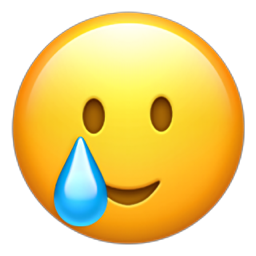}

\newcommand{\wearyface}{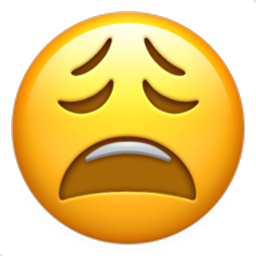}

\newcommand{\hundredpoints}{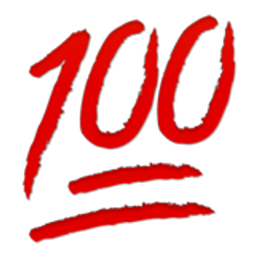}
\newcommand{\facewithrollingeyes}{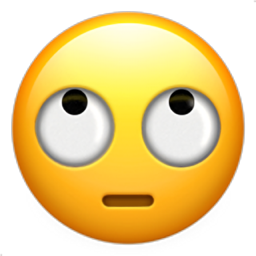}
\newcommand{\blueheart}{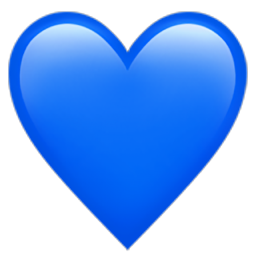}
\newcommand{\purpleheart}{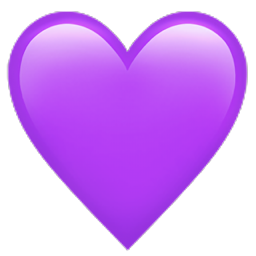}

\newcommand{\flushedface}{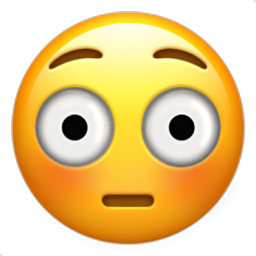}

\newcommand{\shoppingbag}{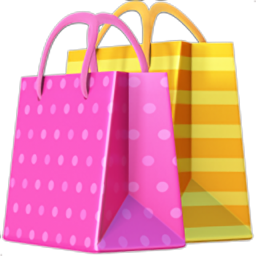}

\newcommand{\nailpolish}{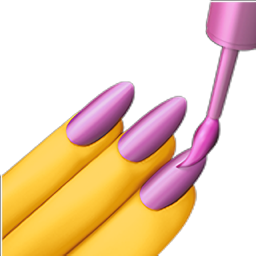}

\newcommand{\womanclothes}{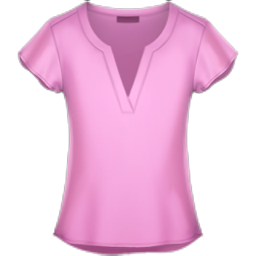}
\newcommand{\purse}{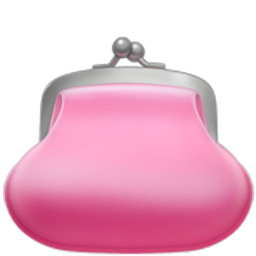}

\newcommand{\trophy}{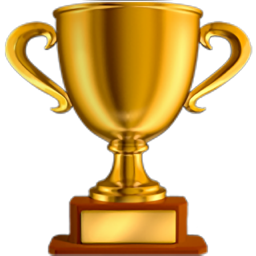}
\newcommand{\sneaker}{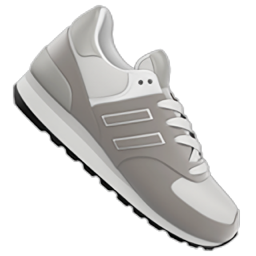}
\newcommand{\cherryblossom}{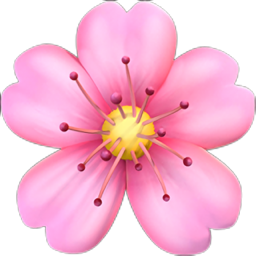}
\newcommand{\balloonemoji}{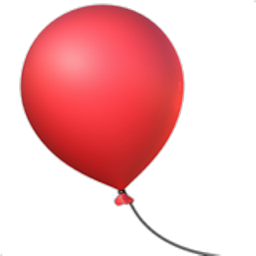}
\newcommand{\astonishedface}{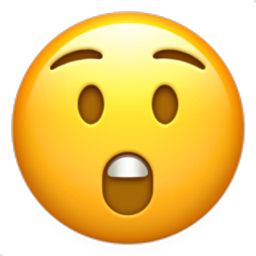}
\newcommand{\birthdaycake}{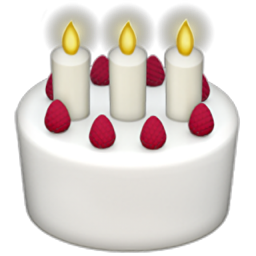}
\newcommand{\cigarette}{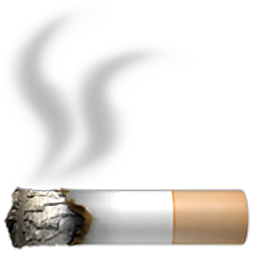}
\newcommand{\malesign}{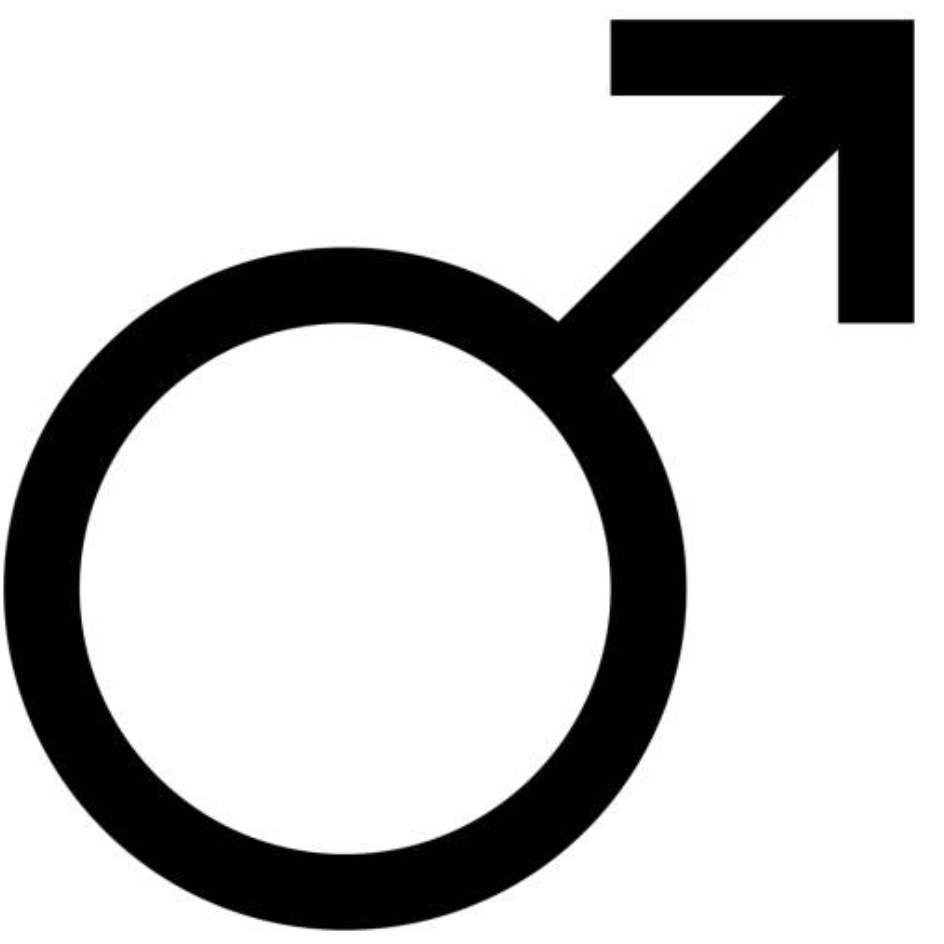}
\newcommand{\palette}{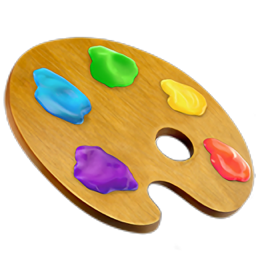}
\newcommand{\pill}{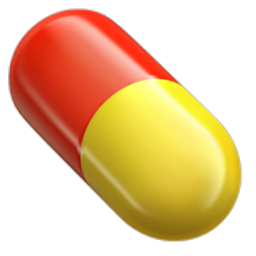}
\newcommand{\broken}{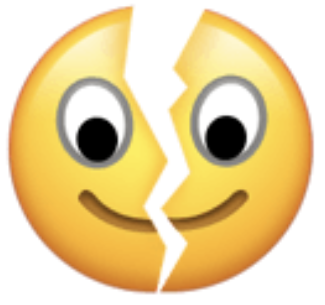}
\newcommand{\onlooker}{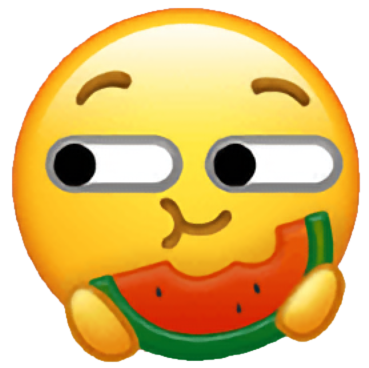}
\newcommand{\doge}{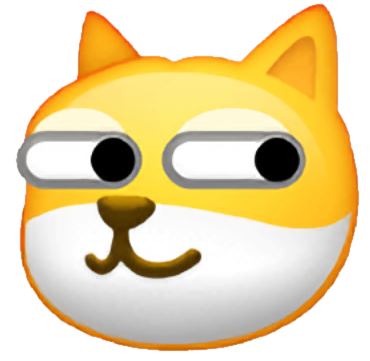}
\newcommand{\cowface}{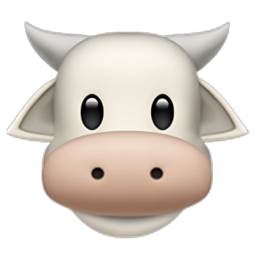}
\newcommand{\beermug}{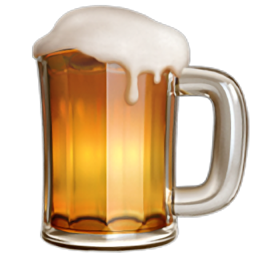}
\newcommand{\sunemoji}{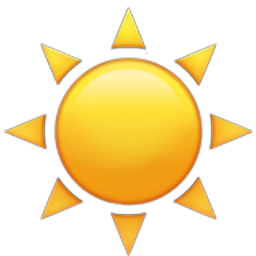}
\newcommand{\usaflag}{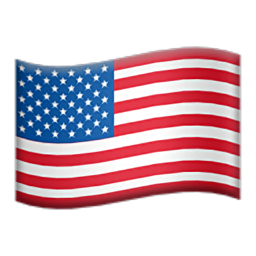}
\newcommand{\cyclist}{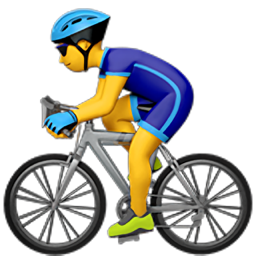}
\newcommand{\rooster}{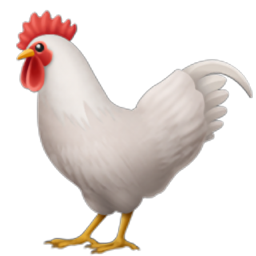}
\newcommand{\womansinger}{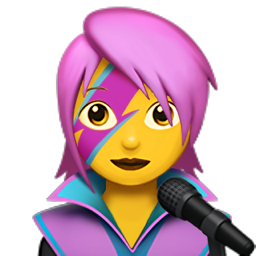}
\newcommand{\highheeled}{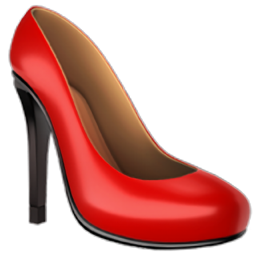}
\newcommand{\ribbon}{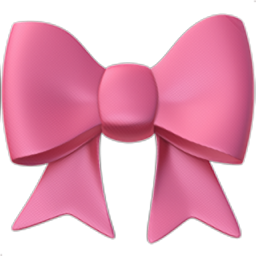}
\newcommand{\maninsuit}{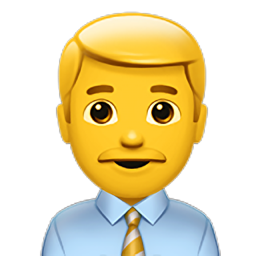}
\newcommand{\football}{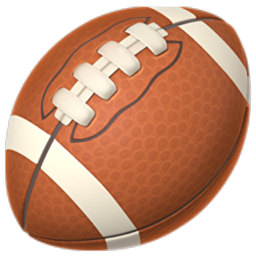}
\newcommand{\automobile}{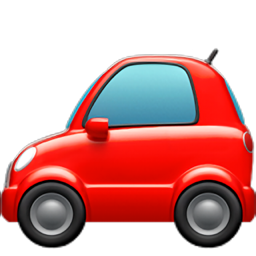}
\newcommand{\hammerwrench}{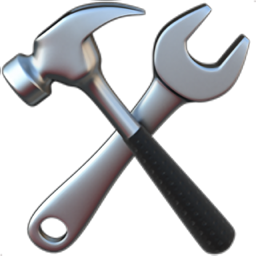}
\newcommand{\cutofmeat}{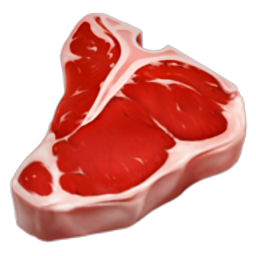}
\newcommand{\selfie}{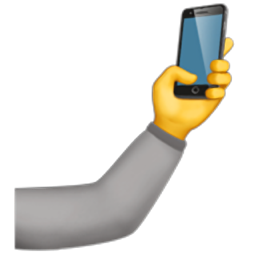}
\newcommand{\breastfeeding}{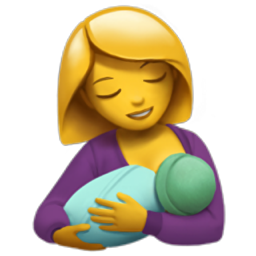}
\newcommand{\confusedface}{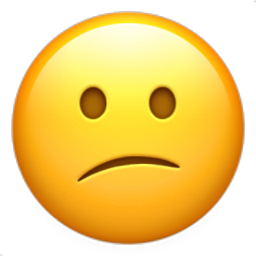}
\newcommand{\memoemoji}{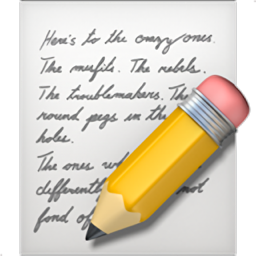}
\newcommand{\packageemoji}{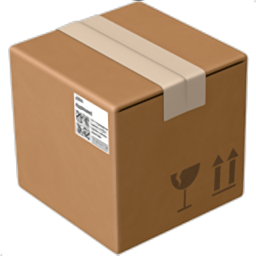}
\newcommand{\globeemoji}{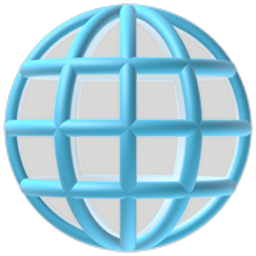}
\newcommand{\lemon}{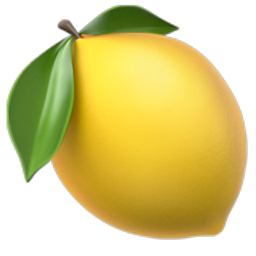}
\newcommand{\melon}{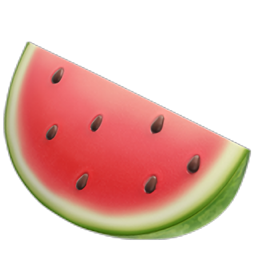}

\newcommand{\bird}{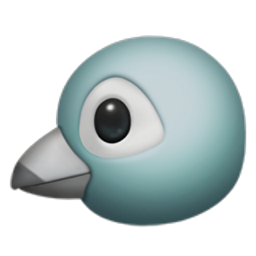}

\usepackage[export]{adjustbox}
\newcommand{\MyEmoji}[1]{\includegraphics[width=1em,valign=t]{#1}}
\newtheorem{observation}{Observation}

\usepackage{newfloat}
\usepackage{listings}
\floatstyle{ruled}
\newfloat{listing}{tb}{lst}{}
\floatname{listing}{Listing}
\title{Emojis Decoded: Leveraging ChatGPT for Enhanced Understanding in Social Media Communications}

\author {
    Yuhang Zhou,\textsuperscript{1}
    Paiheng Xu,\textsuperscript{2}
    Xiyao Wang,\textsuperscript{2}
    Xuan Lu,\textsuperscript{3}
    Ge Gao, \textsuperscript{1}
    Wei Ai \textsuperscript{1} \\
}
\affiliations {
    \textsuperscript{1} College of Information, University of Maryland, College Park, USA \\
    \textsuperscript{2} Department of Computer Science, University of Maryland, College Park, USA \\
    \textsuperscript{3} College of Information Science, University of Arizona, Tucson, USA \\
    \{tonyzhou, paiheng, xywang, gegao, aiwei\}@umd.edu, luxuan@arizona.edu
}

\begin{document}

\maketitle

\begin{abstract}

Emojis, which encapsulate semantics beyond words or phrases, have become prevalent in social network communications. This has spurred increasing scholarly interest in exploring their attributes and functionalities. However, emoji-related research and application face two primary challenges. First, researchers typically rely on crowd-sourcing to annotate emojis in order to understand their sentiments, usage intentions, and semantic meanings. Second, subjective interpretations by users can often lead to misunderstandings of emojis and cause a communication barrier.
Large Language Models (LLMs) have achieved significant success in various annotation tasks, with ChatGPT demonstrating expertise across multiple domains. In our study, we assess ChatGPT's effectiveness in handling previously emoji-annotated and downstream tasks. Our objective is to validate the hypothesis that ChatGPT can serve as an alternative to human annotators in emoji research and that its ability to explain emoji meanings can enhance clarity and transparency in online communications.
Our findings indicate that ChatGPT has extensive knowledge of emojis. It is adept at explaining the meaning of emojis across various application scenarios and demonstrates the potential to replace human annotators in a range of tasks.

\end{abstract}

\section{Introduction}
\label{sec:intro}

Emojis, as prevalent non-verbal tokens in social networks, are crucial in developing an effective discussion in online communication. These language units encode rich semantic meanings that range from emotions, objects, to actions. The crucial role of emojis has attracted a number of researchers to study various emoji-related topics, from emoji functionality, emoji usage patterns, to the emoji application on downstream tasks. 
However, there are two challenges for emoji research and emoji applications for social users. For emoji research, to determine the meaning, sentiments, or intentions of emojis in different scenarios, researchers usually hire human annotators to carry out annotation jobs \cite{lu2018first, hu2017spice}, which is not scalable and time consuming. 
For social users, their differences in demographics, cultural background, or personal experience could lead to misunderstanding of emoji, such as \MyEmoji{\upsidedownface} (upside-down face) to express irony, and incorrect interpretation may cause miscommunication and block emoji diffusion \cite{leonardi2022communication}.


We hypothesize that ChatGPT could offer potential solutions to two key challenges related to emoji interpretation, based on prior explorations.
First, given ChatGPT's demonstrated superiority over human annotators in various text annotation tasks \cite{kocon2023chatgpt}, we believe it may possess a deep understanding of emoji usage. As such, it could replace human workers in labeling emoji attributes, reducing the need for crowd-sourcing efforts and lowering costs.
Second, when users encounter unfamiliar emojis or are unsure of their meanings, ChatGPT’s ability to explain emoji semantics could effectively prevent misunderstandings.

On the other hand, previous work suggests that there exist hallucinations in LLMs' generations \cite{zhang2023siren}, inconsistencies between the model's output and the real-world facts or user input. Whether ChatGPT can produce generations similar to humans for the meaning of emojis is still questionable.
Delineating the boundaries of ChatGPT’s capabilities across emoji-related tasks of varying difficulty is essential for driving progress in future emoji research.

To verify ChatGPT's potential on emoji research and application, our paper conducts a set of qualitative and quantitative studies about ChatGPT's understanding on emoji usage in multiple tasks. Based on the current work on emoji studies, we propose three research questions to probe the ChatGPT's capability on emoji-related tasks.

\begin{description}
    \item[RQ1] Does ChatGPT generate similar explanations on emoji semantics, sentiments, and usage intentions as humans?
    \item[RQ2] Does ChatGPT encode knowledge about emoji usage patterns associated with different communities?
    \item[RQ3] What is the performance of ChatGPT in emoji-related downstream tasks? 
\end{description}

To answer these research questions, we first ask ChatGPT to explain the semantics, sentiments, and intentions of emojis with or without text context, across different platforms or cultures (Section \ref{sec:emoji_semantic}, \ref{sec:emoji_sentiment}, \ref{sec:emoji_intention}). We find that the interpretation of ChatGPT on emojis is consistent with human responses in the existing literature in most cases. Next, we ask ChatGPT to generate the associated emoji patterns with multiple communities (gender, platform, hashtag, and culture) to probe ChatGPT's knowledge about emoji usage in different communities (Section \ref{sec:emoji_community}). 
Finally, we test ChatGPT on emoji-related downstream tasks, i.e., irony annotation and emoji prediction (Section \ref{sec:emoji_irony}, \ref{sec:emoji_recommendation}), and the results reveal that ChatGPT is capable of using the emoji information for irony annotation and making emoji recommendations based on user identity.   
\section{Related Work}
\label{sec:related}
Our work can be related to two streams of previous research studies: emoji functionalities and ChatGPT applications on the computational social science domain.

\subsection{Emoji Interpretation}

With the prevalence of emojis on social networks and other platforms, there has been increasing interest in studying the emoji functionality. Researchers have explored emojis with the functions to express the sentiment, highlight topics, decorate texts, adjust tones, indicate identities, and engage the audience \cite{ai2017untangling, lu2016emojiusgae, ge2019identity, hu2017spice, cramer2016emojifunction}, and with these emoji functions, the researchers summarized the intentions to use emojis on different platforms \cite{hu2017spice, lu2018first}. 
In addition to these functions and intentions, the researchers also noticed differences in the use of emojis in different application scenarios, such as apps, cultures, genders, hashtags, and platforms \cite{tauch2016emojirole, lu2016emojiusgae, chen2018gender, barbieri2016emojiusage, zhou2022emoji, wood2021using}.
However, many works on emoji functionality choose to perform crowd-sourcing work to conduct the emoji interpretation, but with rich and diverse source of emoji applications, it is essential to have an automatic tool to interpret the emoji usage. 
In addition to emoji functionality, several researchers have applied emoji information to downstream tasks, such as emoji prediction\cite{barbieri2020tweeteval, barbieri2018semeval}, improving sentiment analysis \cite{chen2021emoji, chen2019emojipowered, Felbo_2017}, predicting developer's dropout \cite{lu2022emojis}. 

\subsection{LLMs in Computational Social Science}

With the prevalence of LLMs, there are emerging researches evaluating their performances on various tasks \cite{chang2023survey,liu2024large}, including Computational Social Science (CSS) tasks \cite{ziems2023can, zhu2023can}. Researchers have examined ChatGPT’s capacities in the zero-shot or few-shot (with in-context learning) settings, in various text annotation tasks such as political affiliation classification of tweets \cite{tornberg2023chatgpt}; hate speech detection \cite{huang2023chatgpt, li2023hot, zhu2023can}; discourse acts \cite{ostyakova2023chatgpt}; instruction quality evaluation \cite{xu2024promises}; sentiment analysis and bot detection\cite{zhu2023can, zhou2023explore}. A comprehensive evaluation \cite{ziems2023can} shows ChatGPT exhibits unsatisfactory results on tasks that have complex structure or have subjective taxonomies whose semantics differ from definitions learned in pretraining, while achieving better results on tasks that have objective ground truth (e.g., misinformation and fact checking) or have labels with explicit colloquial definitions in the pretraining data (e.g., emotion and political stance).
Nevertheless, the results demonstrate the potential of using LLMs to help researchers with CSS tasks and general users with social-related daily activities.

Despite the growing prominence of emojis as an emerging ``language,'' limited studies focus on the intersection of emojis and ChatGPT. Only a few works suggest ChatGPT's ability to comprehend emojis. \citet{decould2023valence} examines ChatGPT's ability to generate various facial emojis to represent the valence of words, while \citet{peng2023emojilm} explores translating natural language into emoji sequences. \citet{belal2023leveraging} and \citet{kocon2023chatgpt} find that ChatGPT can interpret emojis in texts to enhance sentiment prediction. The most relevant concurrent work is by \citet{lyu2024human}, which investigates whether GPT-4V understands emoji interpretations and correctly employs emojis when given social media posts. However, their study does not cover the functionality or intent of emojis nor discuss emoji usage across different communities.

In contrast, we thoroughly examine the consistency of ChatGPT's annotations on emoji semantics, sentiment, and intentions with human interpretations and probe ChatGPT's knowledge of emoji usage patterns across various communities. We evaluate ChatGPT's performance on emoji-related downstream tasks and explore the significance of emojis to ChatGPT in these tasks. Our evaluation spans multiple languages and platforms, qualitatively and quantitatively demonstrating ChatGPT's understanding of emojis in broader contexts.

\section{Emoji Semantics}
\label{sec:emoji_semantic}



We start evaluating ChatGPT's understanding on emojis by asking ChatGPT to explain the semantic meanings of the emoji. 
Emojis, as universal language tokens, are applied by users with different language and cultural backgrounds. 
In this section, we conduct a qualitative study on the LLM explanation of emoji universal semantics without any given context, and then we probe the ChatGPT understandings in two different language and culture settings: the English context and the Chinese context. 


\begin{figure*}[!h]
     \centering
     \includegraphics[width=\linewidth]{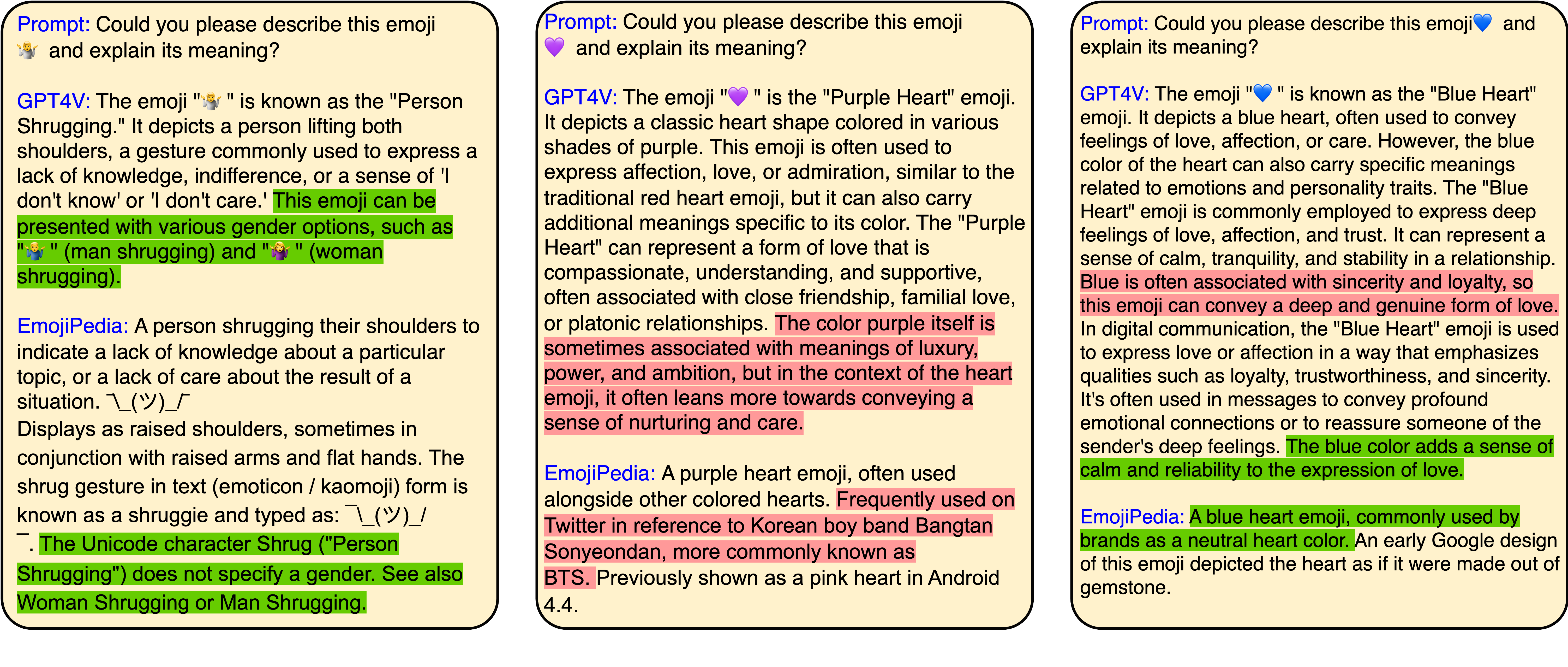}
     \caption{Meaning and explanation of emoji usage from ChatGPT and Emojipedia. Different interpretations are highlighted in red and similar interpretations from both ChatGPT and Emojipedia are highlighted in green.}
     \label{fig:emoji_meaning}
\end{figure*}

\subsection{Emoji Semantics without Context}
\label{sec:emoji_semantics_no_context}

We first collect the 50 emojis most frequently used from 2019 to 2022 provided by the Unicode organization,\footnote{\url{https://home.unicode.org/emoji/emoji-frequency/}} and investigate how ChatGPT explains these emoji semantics without any context. The probing prompt for GPT4 is designed as follows:

\textit{Could you please describe this emoji \texttt{[Emoji]} and explain its meaning?}

where \texttt{[Emoji]} is the Unicode of selected emojis. To compare the understandings between ChatGPT and humans, we also collect the meaning of emojis on the Emojipedia website,\footnote{\url{https://emojipedia.org/meanings}} which is researched and written by Emojipedia editors and lexicographers, as human annotations of emoji meanings. 
Two authors of the paper independently compare the human annotations and ChatGPT responses, finding that ChatGPT and humans provide similar explanations for all 50 emojis. However, we note that both ChatGPT and human annotators highlight points that the other may not cover.

We present three representative examples for the emoji \MyEmoji{\shrugging} (person shrugging), \MyEmoji{\purpleheart} (purple heart), \MyEmoji{\blueheart} (blue heart) in Figure \ref{fig:emoji_meaning}, highlighting the same contents mentioned by humans and ChatGPT by green and different contents by red color. 


From Figure \ref{fig:emoji_meaning}, for the emoji \MyEmoji{\shrugging} (person shrugging), it is consistently observed, both in ChatGPT's responses and human annotations, that this emoji primarily conveys a lack of knowledge without implying any specific gender. Regarding the emoji \MyEmoji{\purpleheart} (purple heart), ChatGPT provides an explanation on the significance of the color purple, while human annotations predominantly discuss its usage in tweets related to a Korean boy band. For the emoji \MyEmoji{\blueheart} (blue heart), ChatGPT exclusively addresses the particular meaning associated with the color blue. However, both ChatGPT and human sources agree that \MyEmoji{\blueheart} typically expresses sentiments of calmness and neutrality.

The findings suggest that the explanation from ChatGPT on semantics of most emojis overlaps with the meaning in Emojipedia. Moreover, the content of ChatGPT and human annotations can be supplementary to each other.

\subsection{Emoji Semantics with Text Context}
\label{sec:emoji_english}
The accurate explanation of ChatGPT in emoji semantics without text context follows our expectation due to the notable performance of ChatGPT in sentiment analysis and tweet-related tasks \cite{zhu2023can, tornberg2023chatgpt}. However, it is well known that emojis will evolve extensive semantics according to the applied text context. For example, \MyEmoji{\goat} (goat) can mean the greatest person, and \MyEmoji{\snake} (snake) can mean the Python language in the specific context. We thus explore whether ChatGPT can interpret an emoji's meaning when giving tweets containing emojis with extensive semantics.
We choose a few emojis (\MyEmoji{\goat} goat, \MyEmoji{\honeybee} honey bee, and \MyEmoji{\snake} snake) with widely accepted special meanings\footnote{selected from this website \url{https://bestlifeonline.com/emoji-meanings/}} and randomly extract tweets in 2022 that contain these emojis via Twitter API. We feed these tweets to ChatGPT and ask ChatGPT to give the semantic explanation based on the context of the tweet. 

\begin{figure*}[!h]
     \centering
     \includegraphics[width=\linewidth]{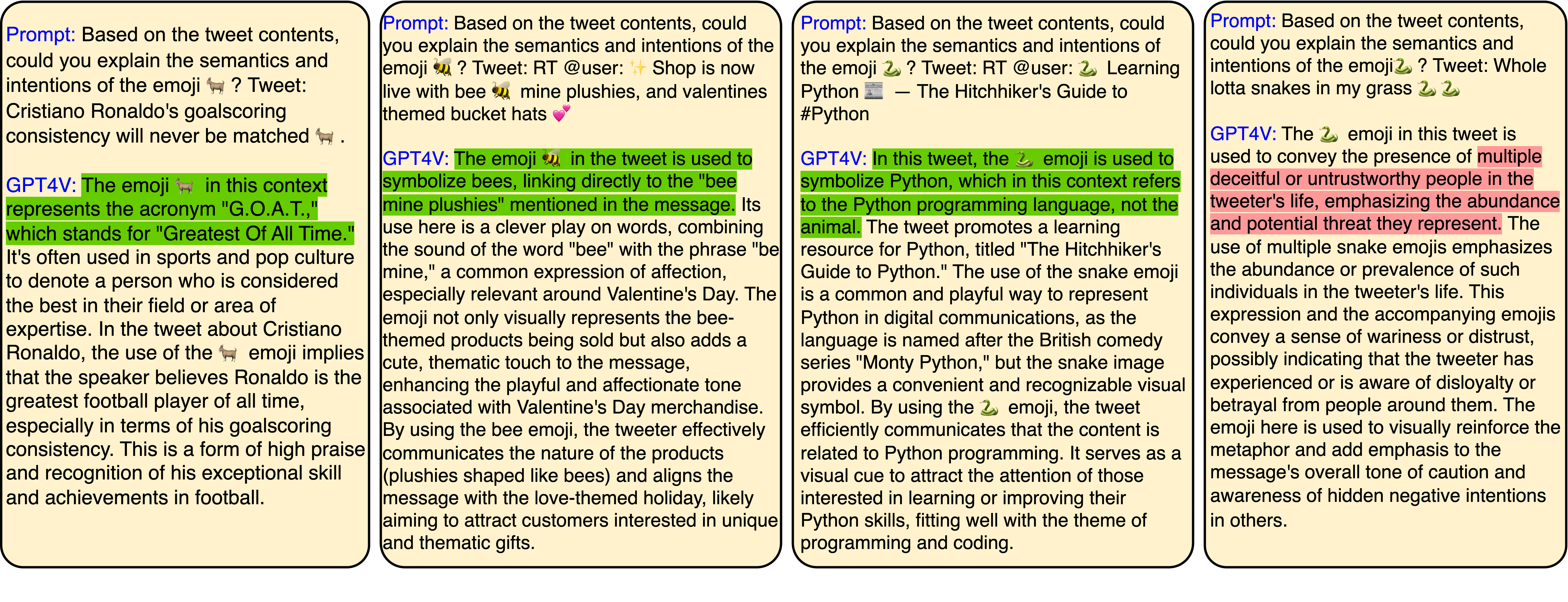}
     \caption{Semantic explanation from ChatGPT of emojis with special meaning. The green color highlights the semantic explanation consistent with human understanding, and the red color highlights the hallucinations. From the case study, ChatGPT can precisely understand the semantics of emojis with special usage when given the specific context in most cases.}
     \label{fig:special_meaning}
\end{figure*}

We present the prompt and the GPT4 response in Figure \ref{fig:special_meaning}, which illustrates that ChatGPT's interpretations of emoji semantics align closely with human understanding in most instances. ChatGPT accurately interprets the emoji \MyEmoji{\goat} (goat) as ``the greatest of all time.'' Additionally, it interprets \MyEmoji{\honeybee} in a Valentine's context as ``be mine.'' 
In the third example featuring the \MyEmoji{\snake} emoji, ChatGPT recognizes its use in symbolizing the Python programming language. However, in the last example with the tweet context ``Whole lotta snakes in my grass \MyEmoji{\snake}\MyEmoji{\snake},'' we argue that the emoji should symbolize the animal snake rather than untrustworthy people. The frequent use of the snake emoji to represent betrayal might have led to this misinterpretation. In conclusion, ChatGPT generally demonstrates a precise understanding of emoji semantics in textual contexts. However, when an emoji represents a less commonly used object, ChatGPT may provide erroneous interpretations.

\subsection{Emoji Semantics with Language/Cultural Context}
\label{sec:semantic_cross_cultural}


In addition to the application in the English context, emojis also show prevalence in other specific language and culture contexts and emojis may evolve to some unique meanings in a specific language context \cite{chen2019emojipowered, lu2016emojiusgae}.
ChatGPT, as a multilingual language model, has demonstrated strong cross-linguistic understanding abilities \cite{openai2023gpt4}. To evaluate the feasibility of ChatGPT replacing human annotators in different language contexts, we select a few emojis with unique meanings in the Chinese context and observe whether ChatGPT could correctly capture the specific semantics.

With a similar experiment setup as in Section \ref{sec:emoji_english}, we first choose emojis that have unique meanings on Chinese social networks and test if ChatGPT can properly explain their semantics, that is, \MyEmoji{\slightsmilingface} (slightly smiling face), \MyEmoji{\doge} (doge), \MyEmoji{\pill} (pill), \MyEmoji{\bugemoji} (bug), \MyEmoji{\lemon}(lemon), \MyEmoji{\melon} (melon), and \MyEmoji{\bird} (bird).
In the Chinese pop culture context, \MyEmoji{\slightsmilingface} ironically expresses dislike and ridicule; \MyEmoji{\doge} conveys a sense of irony or sarcasm; \MyEmoji{\pill} and \MyEmoji{\bugemoji} use a homophone pun, meaning ``it's over'' and expressing support, respectively; \MyEmoji{\lemon} express envy and jealousy; \MyEmoji{\melon} means gossip or drama; \MyEmoji{\bird} refers to not showing up for a planned meeting, when it's used as a pigeon.

Additionally, we include an emoji combo, \MyEmoji{\cowface} \MyEmoji{\beermug}, that consists of a cow face and a beer mug (a homophone pun to express ``so cool''). 
We prompt ChatGPT to explain the usage of these emojis. The prompts are written in English or Chinese, with or without specification of usage in Chinese pop culture. For some emojis, we include appropriate hints, such as connections to homophone puns.

We find that ChatGPT can generate desired interpretations in all cases, but only with appropriate prompts. The prompts need to include very specific information sometimes. For instance, when prompted with ``Explain \texttt{[Emoji]}'', GPT models failed to generate desired explanations and answered with literal interpretations from their Unicode names or visual cues (e.g., \MyEmoji{\slightsmilingface} as friendly instead of its common ironical use in Chinese pop culture \cite{wang2022sarcastic}). When prompted in Chinese, only \MyEmoji{\cowface} \MyEmoji{\beermug} got the correct explanation. However, \MyEmoji{\slightsmilingface}, \MyEmoji{\melon}, and \MyEmoji{\doge} get desired explanations when prompted with ``Explain the \texttt{[Emoji]} in the context of Chinese pop culture'', no matter whether prompted in English or Chinese. 
For \MyEmoji{\pill} and \MyEmoji{\bugemoji}, we had to specifically mention its homophonic usage with y\`ao w\'an and ch\={o}ng, respectively.
Similarly for \MyEmoji{\bird} we had to mention its usage as a pigeon.

We also observe that the Chinese answers generated by prompts written in Chinese contained more specific and useful information in general, compared to the counterparts in English. For example, the English answer for \MyEmoji{\slightsmilingface} mostly conveys that its usage can be context-dependent and culturally specific and only briefly mentions ``it might be used ironically'', while the Chinese version details the several specific contexts when it can be used ironically. 

Meanwhile, answers in both languages are prone to hallucinations even when a large proportion of answers are desired. 
The model links \MyEmoji{\pill} to the red pill and blue pill metaphor from the Matrix movie and incorrectly states that this is a common usage in Chinese social media.

In summary, ChatGPT is capable of correctly explaining the emojis in the Chinese pop culture context. However, prompting in Chinese or mentioning appropriate hints is necessary to better elicit such knowledge in some cases.

\begin{observation}
ChatGPT can output consistent explanations on emoji semantics when giving no context or specific English context. However, when probing the special meaning of an emoji in the Chinese pop culture context, ChatGPT is prone to hallucinations with English prompts but more accurate explanations when prompted in Chinese or given appropriate hints.
\end{observation}

\section{Emoji Sentiment}
\label{sec:emoji_sentiment}

Expressing sentiment is one of the primary functions of emojis, especially face emojis \cite{ai2017untangling}. Previous research has employed human annotation to investigate emoji sentiment \cite{hu2017spice} and has incorporated emoji information to enhance downstream sentiment analysis \cite{Felbo_2017, chen2019emojipowered}. In this section, we prompt ChatGPT to annotate emoji sentiment and compare its annotations with those of humans. Additionally, we evaluate ChatGPT's performance in predicting the sentiment of tweets containing emojis for downstream tasks.

\subsection{Emoji Sentiment Annotation}
\label{sec:sentiment_no_context}
We first ask GPT4 to label the 50 most frequently used emojis in Section \ref{sec:emoji_semantics_no_context} with positive, neutral and negative sentiment with the prompt in Table \ref{tab:sentiment_no_text} in Appendix.

Compared to the human annotations in \citet{hu2017spice}, we observe that ChatGPT and human annotators share the same sentiment annotation for almost all emojis (49 out of 50). The only difference is the annotation for the \MyEmoji{\flushedface} (flushed face) emoji.
While the human annotation is neutral, ChatGPT's response is negative, considering feelings of embarrassment or a sense of being overwhelmed associated with \MyEmoji{\flushedface}. This discrepancy indicates that ChatGPT can interpret the sentiment of emojis but may differ from human annotations in specific cases. ChatGPT's response often notes that it requires text context to accurately determine emoji sentiment. Therefore, we discuss the influence of emojis on ChatGPT's performance in downstream sentiment analysis tasks.

\subsection{Sentiment Prediction with Emojis}
\label{sec:sentiment_prediction}

Previous researchers have emphasized the importance of emoji information to judge sentiment in the given inputs \cite{chen2019emojipowered, Felbo_2017}. We also investigate whether ChatGPT considers emojis in its sentiment classification predictions. For the experiment setup, we choose the sentiment analysis dataset from the TweetEval benchmark \cite{barbieri2020tweeteval}, which includes three labels: negative, neutral, and positive. We extract tweets containing emojis from the dataset, resulting in 805 tweets. We perform sentiment prediction using in-context learning (ICL) with GPT4 and GPT3.5, providing 5 demonstrations for each model. The models achieve 70.31\% and 71.67\% accuracy for GPT4 and GPT3.5, respectively. The fine-tuned RoBERTa model, pretrained on the tweet corpus, achieves 69.3\% accuracy \cite{barbieri2020tweeteval}, indicating that ChatGPT provides reliable sentiment predictions.


Next, we remove the emojis from the input text and ask ChatGPT to assign a sentiment label again. The performance accuracy decreases to 69.56\% and 70.19\% for GPT4 and GPT3.5, respectively. However, this decrease in accuracy does not conclusively demonstrate that ChatGPT utilizes emoji information for sentiment prediction, as the change is not significant and removing emojis may also alter the ground-truth sentiment of the text. We compare the sentences where ChatGPT changes the sentiment prediction after removing the emojis and find that 19.50\% and 18.13\% of tweets flip the prediction labels for GPT3.5 and GPT4 annotations, respectively. We present three randomly selected tweets with flipped predictions for GPT4 in Table \ref{tab:flipped_prediction}.

\begin{table}[!htb]
\centering
\resizebox{\columnwidth}{!}{%
\begin{tabular}{llll}
\toprule
Tweet                                                                                                           & Label     & $p_{emoji}$ & $p_{noemoji}$ \\ \midrule
\begin{tabular}[c]{@{}l@{}}How many more \\ days until opening day? \MyEmoji{\wearyface} \end{tabular}                                & neutral   & negative   & neutral    \\ \midrule
\begin{tabular}[c]{@{}l@{}}@user Everyone knows that vegetarianism \\ is synonymous with wellness! \MyEmoji{\facewithrollingeyes}
\end{tabular} & positive  & neutral    & positive   \\ \midrule
\begin{tabular}[c]{@{}l@{}}\#MPN \#OneDirection\#MtvStarsNiallHoran  \\ JESUS \MyEmoji{\hearteyes}\MyEmoji{\hearteyes}\MyEmoji{\hearteyes}\MyEmoji{\hearteyes}\MyEmoji{\hearteyes}\end{tabular}                       & positive & positive   & neutral    \\ \bottomrule
\end{tabular}
}
\caption{Tweets with flipped sentiment predictions from ChatGPT, where $p_{emoji}$ and $p_{noemoji}$ represents the ChatGPT predictions with and without emojis.}
\label{tab:flipped_prediction}
\end{table}


The analysis of examples in Table \ref{tab:flipped_prediction} reveals that ChatGPT's sentiment predictions are influenced by the presence of emojis within the text. Incorporating emojis like \MyEmoji{\wearyface} (weary face) and \MyEmoji{\facewithrollingeyes} (face with rolling eyes), which carry negative sentiment, results in a shift in sentiment prediction from neutral to negative and from positive to neutral, respectively. Similarly, the addition of multiple \MyEmoji{\hearteyes} (smiling face with heart eyes) emojis leads to an elevation in sentiment prediction from neutral to positive. This case study suggests that ChatGPT considers emoji tokens in text as significant indicators when assessing overall sentiment.

\begin{observation}
GPT-4 annotates emoji sentiment labels similarly to humans and considers emoji sentiment to enhance its predictions when performing sentiment classification tasks.
\end{observation}



\section{Emoji Usage Intention}
\label{sec:emoji_intention}


In addition to sentiment and meaning, previous researchers have also explored the intention behind using emojis, asking human annotators to label emoji intentions with or without context \cite{hu2017spice, lu2018first}. In this section, we follow the approaches and intention definitions from previous work \cite{hu2017spice} and ask ChatGPT to reannotate the emojis with or without context. This allows us to observe whether humans and ChatGPT have similar understandings of the intention behind using emojis.

\subsection{Emoji Intention Annotation without Context}
\label{sec:emoji_intention_wo_context}

First, we annotate emojis without any context to assess ChatGPT's understanding of their universal intentions. We use the same seven types of candidate intentions as in previous work: expressing sentiment, strengthening expression, adjusting tone, expressing humor, expressing irony, expressing intimacy, and describing content. Detailed definitions of these intentions can be found in \citet{hu2017spice}. We select 15 emojis (5 positive, 5 neutral, and 5 negative) from \citet{hu2017spice} and ask ChatGPT to rate the willingness to use each emoji for each intention on a 7-point scale (7 = totally willing, 1 = not willing at all), following the same annotation method as in \citet{hu2017spice}. The complete prompt is shown in Table \ref{table:intention_prompt_no_context} in Appendix.




We calculate the average difference between ChatGPT-rated scores and human ratings as follows: -0.13 (expressing sentiment), -0.29 (strengthening expression), -0.07 (adjusting tone), -0.54 (expressing humor), -1.49 (expressing irony), -0.05 (expressing intimacy), and 1.11 (describing content). Given that the average standard deviation (SD) of human-rated scores is 1.643 \cite{hu2017spice}, ChatGPT and human annotators reach a high consensus on the intentions of expressing sentiment, strengthening expression, and adjusting tone. For other intentions, such as describing content, ChatGPT's scores differ from human ratings, but the discrepancy is smaller than the average human SD. Hence, we conclude that ChatGPT's understanding of the universal usage intention of emojis is generally similar to that of human users.


\subsection{Intention of Emoji Usage in GitHub Posts}



The similar intention annotation without context between ChatGPT and humans aligns with the accurate explanation of emoji semantics and the precise annotation of emoji sentiment. However, the intention behind using emojis can vary across different platforms. For example, users on GitHub may use emojis to organize content within a post but rarely to express irony. Understanding emoji intention in GitHub posts poses a greater challenge for ChatGPT.

In this part, we collect the same 2,000 emoji posts on GitHub with human annotations from \citet{lu2018first} and compare ChatGPT's annotations of emoji intentions with the human ground truth labels. The candidate intentions of emojis in GitHub posts are shown in Table \ref{tab:github_intention_acc}, with detailed definitions provided in \citet{lu2018first}.

\begin{table}[!htb]
\centering
\resizebox{\columnwidth}{!}{%
\begin{tabular}{lllll}
\toprule
Intention & \begin{tabular}[c]{@{}l@{}}Sentiment \\ expressed\end{tabular} & \begin{tabular}[c]{@{}l@{}}Sentiment \\ strengthened\end{tabular} & \begin{tabular}[c]{@{}l@{}}Tone \\ adjusted\end{tabular}         & \begin{tabular}[c]{@{}l@{}}Content \\ described\end{tabular} \\ \midrule
Accuracy  & 67.43\%                                                        & 30.85\%                                                           & 48.64\%                                                          & 47.54\%                                                      \\ \midrule \midrule
Intention & \begin{tabular}[c]{@{}l@{}}Content \\ organized\end{tabular}   & \begin{tabular}[c]{@{}l@{}}Content \\ emphasized\end{tabular}     & \begin{tabular}[c]{@{}l@{}}Non \\ communication use\end{tabular} & Symbol                                                       \\ \midrule
Accuracy  & 40.63\%                                                        & 55.49\%                                                           & 34.62\%                                                          & 14.58\% \\ \bottomrule                                                     
\end{tabular}
}
\caption{Accuracy of ChatGPT annotation for each intention on emojis in GitHub posts with human annotation as the ground truth labels.}
\label{tab:github_intention_acc}
\end{table}


The annotated accuracy between GPT3.5 and GPT4 compared with human annotations is 38.8\% and 49.0\%, respectively. The detailed accuracy of GPT4 for each intention is presented in Table \ref{tab:github_intention_acc}. We observe that for emojis used to express sentiment and emphasize content in GitHub posts, ChatGPT's judgments are more aligned with human annotators. However, for symbols (symbolize emojis) and non-communication use (unintentional use), which are rarely applied on other social platforms, ChatGPT's intention annotations do not match those of human annotators.


We also conduct a detailed case study and present three randomly selected posts with different annotations in Table \ref{tab:github_intention} to further analyze the extent to which ChatGPT's responses align with human understanding.

\begin{table}[!htb]
\centering
\resizebox{\columnwidth}{!}{%
\begin{tabular}{lll}
\toprule
Post                                                                  & Human label     & ChatGPT label \\ \midrule
\begin{tabular}[c]{@{}l@{}l@{}} Make awesome stuff \\ with half the time OR \\ make quadriply \\ awesome creations \\ with the same time \MyEmoji{\sparkles} \end{tabular}  & sentiment strengthened   & content emphasized    \\ \midrule
thanks \MyEmoji{\pray} & sentiment strengthened  &  sentiment expressed   \\ \midrule
\MyEmoji{\clipboard}  Example                       & content organized  & content described    \\ \bottomrule
\end{tabular}
}
\caption{GitHub posts with different annotation labels for emoji usage intention from ChatGPT and humans.}
\label{tab:github_intention}
\end{table}


Furthermore, we feed the ChatGPT annotations in Table \ref{tab:github_intention} back to the LLM and ask for justification. In the first post, human annotators perceive the use of \MyEmoji{\sparkles} as an enhancement of the post's positive sentiment, whereas ChatGPT interprets its use as a means to attract attention. In the second post, which contains the word ``thanks'' accompanied by \MyEmoji{\pray}, human annotators believe that ``thanks'' alone conveys positive sentiment, and the emoji serves to reinforce this sentiment. ChatGPT's interpretation also notices that the emoji conveys the sentiment and gives a similar annotation: sentiment expressed. The final post features \MyEmoji{\clipboard}, which human annotators suggest is used for organizing content to enhance readability, while ChatGPT views it as a symbol or emphasis of a document.
From the case study, we find that even in incorrect annotations, ChatGPT still captures the primary intention behind the emoji usage.
This case study highlights that ChatGPT's interpretations of emoji usage in GitHub posts are not only plausible but also indicative of ChatGPT's nuanced understanding of emoji applications.

\begin{observation}
ChatGPT precisely deduces the intention of emoji usage without context and, when provided with a specific GitHub post, reaches a consensus with human annotations on the intention of emojis.
\end{observation}

\section{Emoji Usage Associated with Communities}
\label{sec:emoji_community}


While ChatGPT has demonstrated a deep understanding of emoji semantics, intentions, and sentiments, previous work suggests that emoji usage can vary significantly across different communities, such as platforms, languages, genders, and hashtags \cite{lu2018first, lu2016emojiusgae, chen2018gender, zhou2022emoji, zhou2023emoji}. In this section, we investigate whether ChatGPT encodes knowledge of these associated emoji usage patterns when provided with prior information about users, platforms, or co-occurring hashtags. We construct the prompt by directly asking for the most used emojis given different community information, and ChatGPT outputs the most used emojis along with its justification.
The probing prompt for ChatGPT is designed as follows:

\textit{Could you output 10 emojis that are associated with the {specific community}?} 

where \textit{{specific community}} refers to the associated community, such as female users, French users, specific hashtags, or the GitHub platform.

We present ChatGPT's answers and compare them with findings from previous work in Tables \ref{tab:emoji_gender}, \ref{tab:emoji_hashtag}, and \ref{tab:emoji_github}. In particular, emojis associated with French, female, and male users are explored in \citet{lu2016emojiusgae, chen2018gender}, and emojis co-occurring with different hashtags and on the GitHub platform are reflected in \citet{zhou2022emoji, zhou2023emoji, lu2018first}.

\begin{table}[!htb]
    \small
    \centering
    \resizebox{\columnwidth}{!}{%
    \begin{tabular}{lcc}
    \toprule
    User         & Emoji from actual usage & Emoji from ChatGPT \\ \midrule
    \begin{tabular}[c]{@{}l@{}l@{}} French user \\ \cite{lu2016emojiusgae} \end{tabular}  & \begin{tabular}[c]{@{}l@{}l@{}} \MyEmoji{\heart} \MyEmoji{\joy} \MyEmoji{\heartwitharrow} \MyEmoji{\sparklingheart} \MyEmoji{\kiss} \\ \MyEmoji{\hearteyes} \MyEmoji{\twohearts} \MyEmoji{\revolvinghearts} \MyEmoji{\purpleheart} \MyEmoji{\growingheart} \end{tabular} & \begin{tabular}[c]{@{}l@{}l@{}} \MyEmoji{\franceflag} \MyEmoji{\wineglass} \MyEmoji{\croissant} \MyEmoji{\tower} \MyEmoji{\kiss} \\ \MyEmoji{\cheese} \MyEmoji{\baguette} \MyEmoji{\palette} \MyEmoji{\clinking} \MyEmoji{\kissmark}  \end{tabular}                        \\ \midrule
    \begin{tabular}[c]{@{}l@{}l@{}} Female user \\ \cite{chen2018gender} \end{tabular}     &  \begin{tabular}[c]{@{}l@{}l@{}} \MyEmoji{\womanholdinghands} \MyEmoji{\birthdaycake} \MyEmoji{\partypopper}  \MyEmoji{\gift} \MyEmoji{\kissmark} \\ \MyEmoji{\balloonemoji} \MyEmoji{\cherryblossom} \MyEmoji{\twohearts} \MyEmoji{\purpleheart} \MyEmoji{\nailpolish} \end{tabular} & \begin{tabular}[c]{@{}l@{}l@{}} \MyEmoji{\womanofficeworker}  \MyEmoji{\womanstudent} \MyEmoji{\womandancing} \MyEmoji{\womansteamy} \MyEmoji{\familywomangirlboy} \\ \MyEmoji{\shoppingbag} \MyEmoji{\nailpolish} \MyEmoji{\womanclothes} \MyEmoji{\purse} \MyEmoji{\cherryblossom} \end{tabular}   \\ \midrule
    \begin{tabular}[c]{@{}l@{}l@{}} Male user \\ \cite{chen2018gender} \end{tabular}   & \MyEmoji{\socceremoji} \MyEmoji{\cigarette} \MyEmoji{\malesign} & \begin{tabular}[c]{@{}l@{}l@{}} \MyEmoji{\smilingfacewithsunglasses} \MyEmoji{\mandancing} \MyEmoji{\flexed} \MyEmoji{\beardman} \MyEmoji{\videogame} \\ \MyEmoji{\socceremoji} \MyEmoji{\hamburger} \MyEmoji{\manlifting} \MyEmoji{\guitar} \MyEmoji{\motorcycle}  \end{tabular}       \\ \bottomrule
    \end{tabular}
    }
    \caption{Outputs from ChatGPT and emojis from actual usage \cite{lu2016emojiusgae, chen2018gender} associated with French users, female users, and male users.}
    \label{tab:emoji_gender}
\end{table}

\begin{table}[!htb]
    \small
    \centering
    \resizebox{\columnwidth}{!}{%
    \begin{tabular}{lcc}
    \toprule
    Hashtag         & Emoji from actual usage & Emoji from ChatGPT \\ \midrule
    \#BlackOutBTS   &  \MyEmoji{\sparkles} \MyEmoji{\relieved} \MyEmoji{\pleadingface} \MyEmoji{\heart} \MyEmoji{\blackheart}                                    &    \MyEmoji{\purpleheart} \MyEmoji{\musicalnote} \MyEmoji{\mandancing} \MyEmoji{\microphone} \MyEmoji{\hearteyes}                            \\ \midrule
    \#MothersDay    &   \MyEmoji{\heart} \MyEmoji{\twohearts} \MyEmoji{\bouquet} \MyEmoji{\whiteheart} \MyEmoji{\dizzy}                                   &                          \MyEmoji{\heart} \MyEmoji {\familywomangirlboy} \MyEmoji{\bouquet} \MyEmoji{\gift} \MyEmoji{\kiss}    \\ \midrule
    \#TheLastDance  &  \MyEmoji{\goat} \MyEmoji{\joy} \MyEmoji{\popcorn} \MyEmoji{\snake} \MyEmoji{\fire}                                    &                                \MyEmoji{\basketball} \MyEmoji{\trophy} \MyEmoji{\TV} \MyEmoji{\sneaker} \MyEmoji{\astonishedface} \\ \bottomrule
    \end{tabular}
    }
    \caption{Outputs from ChatGPT and emojis from actual usage \cite{zhou2022emoji} associated with different hashtags.}
    \label{tab:emoji_hashtag}
\end{table}

\begin{table}[!htb]
    \small
    \centering
    \begin{tabular}{ l  l }
        \toprule
        Source & Emojis associated with GitHub  \\
        \midrule
        \multirow{2}{*}{Actual usage} &  \MyEmoji{\thumbsup} \MyEmoji{\stopsign} \MyEmoji{\smilingeyes} \MyEmoji{\rocket} \MyEmoji{\wavinghand} \MyEmoji{\partypopper} \MyEmoji{\whitecheckmark} \MyEmoji{\bugemoji} \MyEmoji{\warningemoji} \MyEmoji{\thinkingface}  \\
        & \MyEmoji{\sparkles} \MyEmoji{\slightsmilingface} \MyEmoji{\heart} \MyEmoji{\grinningface} \MyEmoji{\ladybugemoji} \MyEmoji{\smilingfacesmilingeyes} \MyEmoji{\moneybag} \MyEmoji{\pray} \MyEmoji{\triangular} \MyEmoji{\bigeyes}\\
        \midrule
        \multirow{2}{*}{ChatGPT} &  \MyEmoji{\thumbsup} \MyEmoji{\thumbsdown} \MyEmoji{\bugemoji} \MyEmoji{\rocket} \MyEmoji{\laptop} \MyEmoji{\partypopper} \MyEmoji{\penguin} \MyEmoji{\sparkles} \MyEmoji{\books} \MyEmoji{\hammer}  \\
        & \MyEmoji{\counterclockwisearrowsbutton} \MyEmoji{\thinkingface} \MyEmoji{\globalemoji} \MyEmoji{\fire} \MyEmoji{\construction} \MyEmoji{\warningemoji} \MyEmoji{\snake} \MyEmoji{\chart} \MyEmoji{\robot} \MyEmoji{\calendaremoji}\\
        \bottomrule
    \end{tabular}
    \caption{Outputs from ChatGPT and emojis from actual usage \cite{zhou2023emoji} associated with the GitHub platform.}
    \label{tab:emoji_github}
\end{table}


From Tables \ref{tab:emoji_gender}, \ref{tab:emoji_hashtag}, and \ref{tab:emoji_github}, we observe that ChatGPT relies on stereotypical associations with user attributes or platform characteristics to select emojis. In most cases, this stereotypical selection results in choosing the same or semantically similar emojis to those identified in previous works.

For example, both ChatGPT and previous works mention \MyEmoji{\socceremoji} (soccer) and \MyEmoji{\cherryblossom} (cherry blossom) as emojis associated with male and female users, respectively. Regarding the most frequently used emojis on the GitHub platform, 7 emojis overlap between previous findings and ChatGPT's answers. Additionally, the other emojis from ChatGPT also reflect attributes of the GitHub platform, such as \MyEmoji{\penguin} (penguin) for Linux and \MyEmoji{\snake} (snake) for Python.


However, since ChatGPT utilizes stereotypical associations to infer associated emojis, the results may sometimes be incomplete. For example, while previous research indicates that French users often use heart-related emojis \cite{lu2016emojiusgae}, this is not reflected in ChatGPT's responses. Similarly, although ChatGPT uses basketball emojis to indicate \#TheLastDance about Michael Jordan, it does not include \MyEmoji{\goat} (the greatest of all time), which often co-occurs with references to Michael Jordan. In summary, the results indicate that ChatGPT notices differences in emoji usage across different contexts, but it cannot cover all facts or details in the emoji domain.

\begin{observation}
ChatGPT employs stereotypical associations to infer emoji usage patterns associated with distinct communities, accurately reflecting variations in emoji use across genders, hashtags, and platforms. However, it fails to capture the nuances of emoji usage specific to different countries. 
\end{observation}

\section{Irony Annotation with Emojis}
\label{sec:emoji_irony}



In Section \ref{sec:emoji_intention}, ChatGPT has reached a consensus with humans about the functionality of emojis to express irony. To assess ChatGPT's proficiency in discerning ironic emoji usage within distinct linguistic contexts, we initiated an ICL experiment on two datasets. One dataset comprises Arabic texts with annotation labels on the irony of emojis. The other consists of English tweets with emojis and annotation labels on the irony of the overall tweet.

For the English dataset, we collected 405 tweets with emojis from the training dataset SemEval2018 Task 3 Subtask A \cite{van2018semeval}, which includes binary annotations indicating whether each tweet is ironic or not. Among the 405 tweets collected, 175 are annotated by humans as ironic, and the remaining are not.

We first used ChatGPT to predict the irony in the original tweets with emojis via ICL inference. We then asked ChatGPT to annotate the tweets again, this time excluding the emojis. The prompt details are provided in Table \ref{table:tweet_irony}. For tweets with emojis, the accuracy of irony prediction with GPT4 is 81.2\%. When the emojis are excluded, the accuracy decreases to 77.0\%. Note that the accuracy drop does not necessarily mean that ChatGPT uses emojis to make irony predictions, as the irony of a sentence may change when emojis are removed. We conducted a case study and present three randomly selected ironic tweets with flipped predictions after removing emojis in Table \ref{tab:irony_flipped_prediction}.

\begin{table}[!htb]
\centering
\resizebox{\columnwidth}{!}{%
\begin{tabular}{llll}
\toprule
Tweet                                                                                                           & Label     & $p_{emoji}$ & $p_{noemoji}$ \\ \midrule
work should be fun today \MyEmoji{\unamusedface}                                & irony   & irony   &  no irony    \\ \midrule
So this week is just getting better and better \MyEmoji{\sob} & irony  & irony    & no irony   \\ \midrule
\begin{tabular}[c]{@{}l@{}}I have three test and a two dance performances \\ tomorrow!! \MyEmoji{\books}\MyEmoji{\openbook}\MyEmoji{\womandancing}\MyEmoji{\womandancing}\MyEmoji{\peoplewithbunnyears} \#EasyDay\end{tabular}                       & irony & no irony   & irony   \\ \bottomrule
\end{tabular}
}
\caption{Tweets with flipped irony predictions from ChatGPT, where $p_{emoji}$ and $p_{noemoji}$ represents the ChatGPT predictions with and without emojis.}
\label{tab:irony_flipped_prediction}
\end{table}


In the analysis of the first two tweets from Table \ref{tab:irony_flipped_prediction}, we observe a shift in irony prediction from ``ironic'' to ``non-ironic'' when emojis are excluded. This shift is attributable to the irony inherent in the emojis \MyEmoji{\unamusedface} and \MyEmoji{\sob}. The textual content of these tweets conveys positive sentiment, but the juxtaposition with negative emojis creates an ironic tone. ChatGPT effectively recognizes the discordance between the sentiments expressed in the text and the emojis, assigning an irony label. However, once the emojis are removed, this contradiction vanishes, leading ChatGPT to revise its classification to ``non-ironic''. In the final tweet, the irony is embedded in the hashtag \#EasyDay, contrasting with the ostensibly busy scenario described in the text. The presence of emojis like \MyEmoji{\books}, \MyEmoji{\womandancing}, and \MyEmoji{\peoplewithbunnyears}, which do not inherently convey irony, potentially obscures ChatGPT's ability to discern the ironic intent. Removing these emojis enables ChatGPT to accurately detect the irony in the tweet. Overall, this case study illustrates that ChatGPT considers emoji tokens as crucial elements in assessing irony in English tweets, demonstrating its nuanced understanding of emojis' role in conveying irony.


To evaluate ChatGPT's understanding of irony in emojis for other languages, we utilize a recently published Arabic dataset with ironic annotations: ArSarcasMoji from \citet{hakami2023arsarcasmoji}. Instead of labeling the irony of sentences, ArSarcasMoji provides the ground truth irony label of emojis in three categories: sarcastic, humorous, and not ironic. The sarcastic label indicates that ironic emojis play a negative sentiment role, while the humorous label indicates a positive role \cite{hakami2023arsarcasmoji}.

Since the label distribution in the original dataset is biased, we first downsample sentences labeled as ``not ironic'' and ``sarcastic'' to create a balanced dataset for evaluation. We compose the prompt with instructions, detailed definitions of ``humorous'' and ``sarcastic,'' and examples of Arabic sentences with labels (the complete prompt is provided in Table \ref{table:irony_arabic}). The prediction accuracy for GPT4 and GPT3.5 is 60.1\% and 63.1\%, respectively, indicating that ChatGPT can understand the functionality of emojis to express irony in the Arabic context.

\begin{observation}
In the emoji irony classification task, ChatGPT demonstrates promising performance in distinguishing whether an emoji is ironic or not. In the tweet irony classification task, ChatGPT considers the presence of emojis when determining the overall irony of a tweet.
\end{observation}

\section{Emoji Recommendation}
\label{sec:emoji_recommendation}

Emoji recommendation, a well-defined and widely adopted task, is essential for evaluating models' understanding of emoji usage \cite{barbieri2020tweeteval, lyu2024human}. This task requires models to grasp both the semantics of emojis and their application context. After establishing that ChatGPT can identify the sentiments, meanings, and intentions behind emojis, we examine its ability to recommend suitable emojis based on a given text context.

\subsection{Recommendation Based on Text Context}



For the experiment, we use the SemEval 2018 Task 2 dataset for emoji recommendation in English and Spanish \cite{barbieri2018semeval}. We select the validation sets with 5,000 English and Spanish tweets for inference. The task pre-defines a list of 20 emojis for English and 19 for Spanish, and our prompt includes the predefined emoji candidates and five exemplars from the respective training set. The prompt details are shown in Table \ref{table:emoji_prediction}.

For emoji recommendation on English tweets, GPT4 and GPT3.5 achieve 22.12\% and 21.52\% accuracy, respectively, while the fine-tuned FastText classifier achieves 42.56\% accuracy. For Spanish tweets, GPT4 and GPT3.5 obtain 20.14\% and 20.96\% accuracy, respectively, while the fine-tuned FastText classifier achieves 31.63\% accuracy. These results suggest that, in a closed-set emoji recommendation, the ICL of ChatGPT cannot outperform the traditional fine-tuned classifier. However, ChatGPT's performance is consistent across languages, whereas the fine-tuned classifier experiences a significant performance drop. Furthermore, we observe ChatGPT hallucinations during emoji recommendation, with 6.1\% and 3.7\% of English tweets being annotated with emojis not in the predefined candidates for GPT3.5 and GPT4, respectively.

\subsection{Recommendation with User Identity}


Previous work suggests that emoji preferences differ based on user identity \cite{chen2018gender, lu2016emojiusgae}, so it is important to consider user demographic attributes when recommending emojis. ChatGPT has shown the potential to act as a communicative agent with different roles \cite{li2023camel}. Moreover, as shown in Section \ref{sec:emoji_community}, ChatGPT encodes knowledge about discrepancies in emoji usage patterns among different communities. In this part, we investigate whether ChatGPT can consider user identity when recommending emojis for tweets.

We ask ChatGPT to act as a female or male Twitter user and provide emoji recommendations accordingly. We add another sentence at the beginning of the prompt, as in previous work \cite{li2023camel}: \textit{I want you to act as a female/male Twitter user. Never forget your role.} We repeat the GPT4 ICL experiment on the same 5,000 English tweets. Acting as a male or female Twitter user, ChatGPT achieves 21.04\% and 20.94\% accuracy, respectively, with 1,110 tweets having different predictions.
Although the provided gender identity does not significantly change the accuracy, it altered more than 20\% of the predictions of the recommended emojis.
We also performed a Chi-Square Test \cite{rana2015chi} to examine the significance of the difference in emoji distributions between the two prompts, yielding a p-value of less than 0.01.
We present the distribution of the top 10 predicted emojis for female and male users in the first two rows of Table \ref{tab:emoji_dist}.

\begin{table}[!htb]
\centering
\resizebox{\columnwidth}{!}{%
\begin{tabular}{ll}
\toprule
Role     & Emoji distribution                                                                                                                \\ \midrule
Female   & \begin{tabular}[c]{@{}l@{}}\MyEmoji{\sparkles}: 657, \MyEmoji{\smilingfacewithsmilingeyes}: 582, \MyEmoji{\smilingfacewithhearts}: 573, \MyEmoji{\fire}: 418, \MyEmoji{\twohearts}: 410, \\ \MyEmoji{\smilingfacewithsunglasses}: 367, \MyEmoji{\joy}: 347, \MyEmoji{\stuckouttonguewinkingface}: 224, \MyEmoji{\christmastree}: 194, \MyEmoji{\heart}: 186\end{tabular} \\ \midrule
Male     &  \begin{tabular}[c]{@{}l@{}}\MyEmoji{\sparkles}: 575, \MyEmoji{\smilingfacewithsmilingeyes}: 560, \MyEmoji{\smilingfacewithhearts}: 527, \MyEmoji{\smilingfacewithsunglasses}: 488, \MyEmoji{\fire}: 451, \\ \MyEmoji{\joy}: 341, \MyEmoji{\heart}: 237, \MyEmoji{\blueheart}: 227, \MyEmoji{\twohearts}: 219, \MyEmoji{\stuckouttonguewinkingface}: 208\end{tabular}                                                                                                                                 \\ \midrule
French   &  \begin{tabular}[c]{@{}l@{}}\MyEmoji{\smilingfacewithsmilingeyes}: 620, \MyEmoji{\smilingfacewithhearts}: 583, \MyEmoji{\sparkles}: 530, \MyEmoji{\smilingfacewithsunglasses}: 411, \MyEmoji{\joy}: 396, \\ \MyEmoji{\fire}: 381, \MyEmoji{\twohearts}: 310 \MyEmoji{\heart}: 252, \MyEmoji{\stuckouttonguewinkingface}: 236, \MyEmoji{\camerawithflash}: 192\end{tabular} \\ \midrule
American &   \begin{tabular}[c]{@{}l@{}}\MyEmoji{\smilingfacewithsmilingeyes}: 605, \MyEmoji{\smilingfacewithhearts}: 576, \MyEmoji{\sparkles}: 566, \MyEmoji{\fire}: 448, \MyEmoji{\smilingfacewithsunglasses}: 408, \\ \MyEmoji{\joy}: 361, \MyEmoji{\twohearts}: 270, \MyEmoji{\stuckouttonguewinkingface}: 233, \MyEmoji{\heart}: 216, \MyEmoji{\christmastree}: 207\end{tabular}                                                                                                                                 \\ \bottomrule
\end{tabular}
}
\caption{Distributions of top 10 predicted emojis when asking ChatGPT to play different roles. The yellow color highlights the emojis \MyEmoji{\twohearts} and \MyEmoji{\smilingfacewithsunglasses}, which are two emojis with the largest change in the number of recommendations for acting as male or female users.}
\label{tab:emoji_dist}
\end{table}

An examination of the first two rows of Table \ref{tab:emoji_dist} reveals a comparable emoji distribution with two notable exceptions: a higher frequency of \MyEmoji{\twohearts} (two hearts) when representing female users, and a higher usage of \MyEmoji{\smilingfacewithsunglasses} (smiling face with sunglasses) in representations of male users. This finding is consistent with ChatGPT's understanding of usage differences for female and male users in Section \ref{sec:emoji_community} and the findings from previous work \cite{chen2018gender}.

To further explore ChatGPT's comprehension of variations in emoji usage across different countries, an aspect not revealed by the results in Section \ref{sec:emoji_community}, we conducted additional experiments. In these, we simulate interactions while portraying users from France and the United States, respectively. When asking ChatGPT to act as a French user, we translate the prompts and tweets to French. We present the predicted emoji distribution in the last two rows of Table \ref{tab:emoji_dist}.

From the last two rows, although the p-value of Chi-Square Test is still smaller than 0.01, we cannot find a significant change in the occurrence numbers for a specific emoji when portraying users from France or the USA. In conjunction with the findings presented in Section \ref{sec:emoji_community} regarding ChatGPT's interpretation of emoji usage by French users, it can be concluded that ChatGPT lacks awareness of the variations in emoji usage across different countries.

\begin{figure*}[!htb]
    \centering
    \includegraphics[width=1\linewidth]{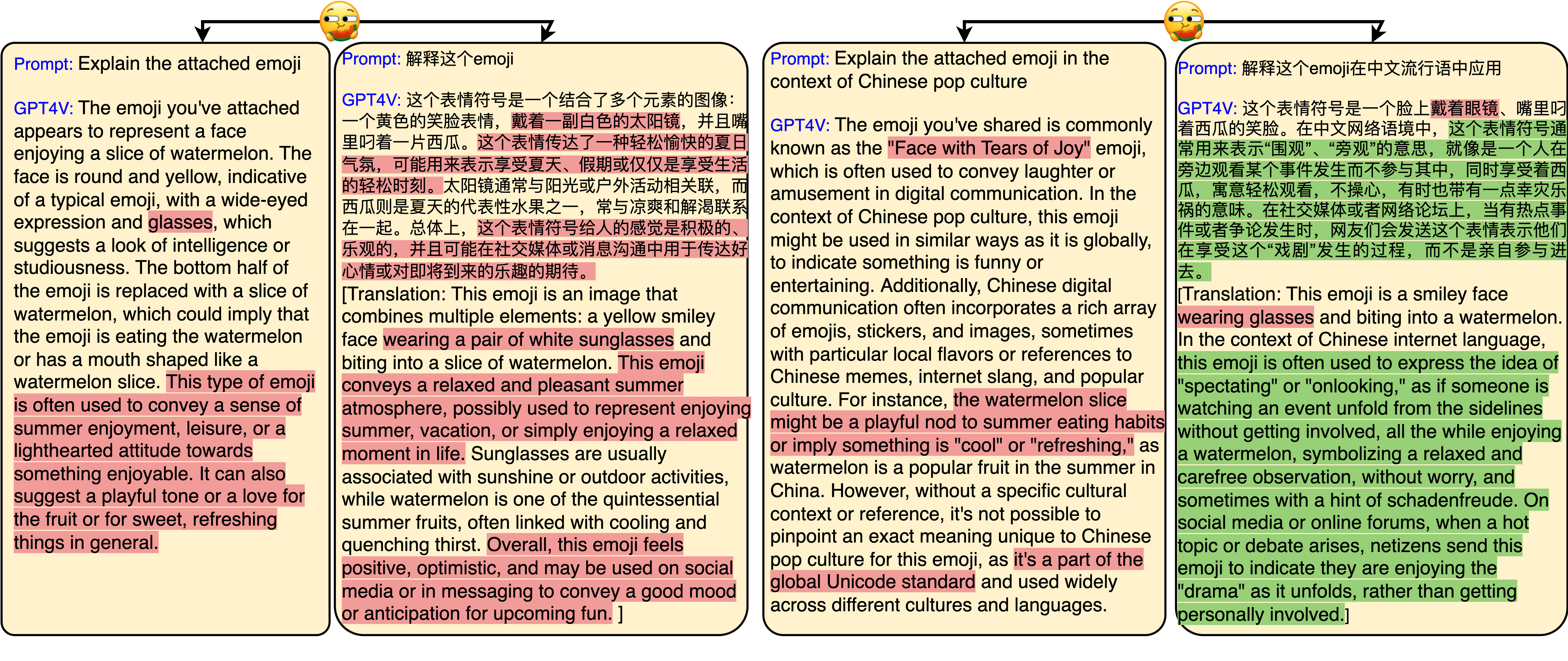}
    \caption{Qualitative Study of Emoji Semantics with
Language/Cultural Context. Hallucinations are highlighted in red and desired interpretations are highlighted in green. }
    \label{fig:cross_lingual}
\end{figure*}

\begin{observation}
ChatGPT does not outperform the fine-tuned traditional classifier on the closed-set emoji recommendation task. However, when provided with user gender information, ChatGPT can alter its emoji recommendations.
\end{observation}
\section{Implications and Limitations}
\label{sec:limitation}


Our paper reveals ChatGPT's precise understanding of emoji functionality and intention. ChatGPT's capabilities with emojis could benefit researchers and social media users. For researchers involved in future emoji studies, leveraging human-aligned explanations of emoji semantics and intentions could significantly reduce the need for extensive crowd-sourcing, particularly in multilingual contexts. This approach would enable more effective and efficient research into emoji usage and interpretation. Social media users could utilize ChatGPT to decipher the semantics of unfamiliar emojis, thereby enhancing the clarity and transparency of online communication.

In addition to ChatGPT's overall effectiveness in emoji-related tasks, we observed instances of hallucination. For example, when explaining the semantics of the \MyEmoji{\snake} emoji in Figure \ref{fig:special_meaning}, emojis with unique meanings on Chinese social media, or when identifying frequently used emojis by French users in Table \ref{tab:emoji_gender}, ChatGPT often relies on spurious correlations or prototypes formed during training or alignment, leading to hallucinations. These findings are valuable for future emoji research, reminding researchers to exercise caution regarding ChatGPT's hallucinations, especially in cross-cultural contexts. They also highlight the need for model developers to focus more on improving non-verbal token understanding in cross-cultural settings.

All the experiments in our paper treat emojis as text tokens in the Unicode form. But some emojis due to its application in the specific cultural context or new appearance are not included in the Unicode organization, such as \MyEmoji{\broken} (broken) and \MyEmoji{\onlooker} (onlooker) in Chinese social media. For these emojis without standard Unicode, we can utilize the vision language model (VLM) from ChatGPT: GPT4V, which has been demonstrated with significant progress in image understanding \cite{yang2023dawn, wang2024mementos}, to help us understand the emoji functions.
We prompt the images of \MyEmoji{\broken} and \MyEmoji{\onlooker} to GPT4V for semantic explanations. An example is shown in Figure \ref{fig:cross_lingual}. We observe similar patterns as in Section \ref{sec:semantic_cross_cultural}. GPT4V generates hallucinations with Chinese prompts but outputs the correct explanation when supplementing the prompt with the context of Chinese pop culture.

We also repeat the annotation of the sentiment of emojis in Section \ref{sec:emoji_semantics_no_context} on GPT4V. For the 50 selected emojis, only \MyEmoji{\pleadingface} (pleading face) and \MyEmoji{\eyes} (eyes) have different annotations for GPT4 and GPT4v. 
For the \MyEmoji{\pleadingface} emoji, VLM attributes a negative sentiment based on its visual features: frown, wide eyes, and raised eyebrows. In contrast, GPT4 associates the \MyEmoji{\pleadingface} emoji with positive sentiment, perceiving it as an expression of vulnerability, empathy, and affection. Regarding the \MyEmoji{\eyes} emoji, GPT4 assigns a neutral sentiment, indicative of observation, while GPT4v leans towards a neutral, yet slightly positive, reflecting an underlying connotation of interest. 

These findings suggest that, while the GPT4v and GPT4 sentiment annotations are generally the same, divergent interpretations of the \MyEmoji{\pleadingface} emoji can highlight the evolution of the emoji sentiment derived from the user application setting. The results in GPT4V imply that, when encountering specific emojis without the standard Unicode, the notable capability in emoji understanding of GPT4V can also help researchers and users. 

With the rapid evolution of current large language models (LLMs), there may be concerns about whether the observations and conclusions presented in this study will still hold for future versions of ChatGPT. On the one hand, in most scenarios where ChatGPT produces human-aligned responses, we anticipate that its accuracy will improve as the model continues to evolve. On the other hand, we also identified cases where ChatGPT produces hallucinations, particularly in the cross-cultural contexts discussed in Sections \ref{sec:emoji_community} and \ref{sec:semantic_cross_cultural}. Since knowledge bases related to platforms like WeChat and Twitter are either not publicly available or restricted for commercial use, we infer that these hallucinations may arise from a lack of relevant knowledge in the training data. We argue that these issues may not be easily resolved by simply scaling up data or model size. Consequently, we expect that the current limitations we have observed are likely to persist and may not be quickly addressed. Therefore, our findings, which delineate the boundaries of ChatGPT’s capabilities, may help researchers better utilize the tool for emoji annotation tasks over time.

There are several limitations in our exploration work. First, previous work has shown that the responses of LLMs are sensitive to temperature and the prompt \cite{yoo2021gpt3mix, schick2020exploiting, gilardi2023chatgpt, zhou2023scalable}, but in our work, for each task, we only explore one type of prompt presented in Appendix. 
For the quantitative studies in our paper, such as intention prediction, irony prediction, sentiment classification, and emoji recommendation, we use the ChatGPT API (gpt-4-1106-preview and gpt-3.5-turbo-1106) with a temperature setting of 0 to generate the model outputs. For the qualitative studies (emoji semantics and emoji usage within communities), we initially used the ChatGPT web interface to obtain results. To demonstrate the robustness of our findings, we repeated qualitative studies using the ChatGPT API with a temperature of 0 and presented the results in the Appendix \ref{sec:chatgpt_api}.
Additionally, to gain further insights into open-source models, we replicated some experiments using Llama3-8B with a temperature setting of 0, and the results are also provided in the Appendix \ref{sec:llama3}. We found that the open-source Llama3-8B model exhibits significant hallucinations when probing emoji intentions or probing the associations between emoji and community information, making it unreliable for replacing human annotators with the base Llama3-8B model.

Second, emoji semantics or sentiment can evolve over time. For example, \MyEmoji{\slightsmilingface} (slightly smiling face) has evolved the sense of irony and negative sentiment in Chinese social networks. The human annotations utilized in this paper are almost all prepared before 2018, and the 5-year gap may lead to the different annotation results.

Lastly, our paper focuses on the ICL of GPT3.5, GPT4 and GPT4V for ChatGPT and does not involve the fine-tuning experiment with OpenAI API. Moreover, with the release of GPTs (the tailored version of ChatGPT for a specific task \footnote{\url{https://openai.com/blog/introducing-gpts}}), we can find there is a version of ChatGPT from OpenAI, called \textit{genz 4 meme}, to help understand the latest memes, which may have better understanding for emojis. We leave further exploration of the different versions of ChatGPT in the future work.

\section{Conclusion}

In this study, we conduct a comprehensive evaluation of ChatGPT's proficiency in interpreting emojis, focusing on various aspects including meaning, sentiment, and intention. Our analysis reveals that ChatGPT's annotations closely align with human labels. We discover that ChatGPT has embedded knowledge of emoji usage patterns prevalent in different communities. Furthermore, ChatGPT exhibits strong performance in several downstream tasks that involve emojis, indicating its capability to leverage its emoji understanding in making predictions. The encouraging results in annotation tasks suggest that ChatGPT could significantly contribute to emoji research by serving as a substitute for human annotators, thus conserving human resources.



\bibliography{abbrev, aaai22}

\appendix

\section*{Paper Checklist}
\begin{enumerate}

\item For most authors...
\begin{enumerate}
    \item  Would answering this research question advance science without violating social contracts, such as violating privacy norms, perpetuating unfair profiling, exacerbating the socio-economic divide, or implying disrespect to societies or cultures?
    \answerYes{Yes}
  \item Do your main claims in the abstract and introduction accurately reflect the paper's contributions and scope?
    \answerYes{Yes, see the Abstract and the Introduction}
   \item Do you clarify how the proposed methodological approach is appropriate for the claims made? 
    \answerYes{Yes, see Section 3 to 8}
   \item Do you clarify what are possible artifacts in the data used, given population-specific distributions?
    \answerYes{Yes, see Section 3 to 8}
  \item Did you describe the limitations of your work?
    \answerYes{Yes, see the Limitations and Implications}
  \item Did you discuss any potential negative societal impacts of your work?
    \answerYes{Yes, see the Limitations and Implications}
      \item Did you discuss any potential misuse of your work?
    \answerYes{Yes, see the Limitations and Implications}
    \item Did you describe steps taken to prevent or mitigate potential negative outcomes of the research, such as data and model documentation, data anonymization, responsible release, access control, and the reproducibility of findings?
    \answerYes{Yes, see the Limitations and Implications}
  \item Have you read the ethics review guidelines and ensured that your paper conforms to them?
    \answerYes{Yes}
\end{enumerate}

\item Additionally, if your study involves hypotheses testing...
\begin{enumerate}
  \item Did you clearly state the assumptions underlying all theoretical results?
    \answerNA{NA}
  \item Have you provided justifications for all theoretical results?
    \answerNA{NA}
  \item Did you discuss competing hypotheses or theories that might challenge or complement your theoretical results?
    \answerNA{NA}
  \item Have you considered alternative mechanisms or explanations that might account for the same outcomes observed in your study?
    \answerNA{NA}
  \item Did you address potential biases or limitations in your theoretical framework?
    \answerNA{NA}
  \item Have you related your theoretical results to the existing literature in social science?
    \answerNA{NA}
  \item Did you discuss the implications of your theoretical results for policy, practice, or further research in the social science domain?
    \answerNA{NA}
\end{enumerate}

\item Additionally, if you are including theoretical proofs...
\begin{enumerate}
  \item Did you state the full set of assumptions of all theoretical results?
    \answerNA{NA}
	\item Did you include complete proofs of all theoretical results?
    \answerNA{NA}
\end{enumerate}

\item Additionally, if you ran machine learning experiments...
\begin{enumerate}
  \item Did you include the code, data, and instructions needed to reproduce the main experimental results (either in the supplemental material or as a URL)?
    \answerYes{Yes, see Section 3 to 8. Our experiment result is all about the in-context learning of ChatGPT, which only needs to call OpenAI API to get the results.}
  \item Did you specify all the training details (e.g., data splits, hyperparameters, how they were chosen)?
    \answerYes{Yes, the prompt details are in Appendix Section}
     \item Did you report error bars (e.g., with respect to the random seed after running experiments multiple times)?
    \answerNo{No, we fix the temperature to 0, so the results are fixed}
	\item Did you include the total amount of compute and the type of resources used (e.g., type of GPUs, internal cluster, or cloud provider)?
    \answerNo{No, calling OpenAI API does not cost computation resources}
     \item Do you justify how the proposed evaluation is sufficient and appropriate to the claims made? 
    \answerNo{No, for quantitative results, we use accuracy as the evaluation metric}
     \item Do you discuss what is ``the cost`` of misclassification and fault (in)tolerance?
    \answerNA{NA}
  
\end{enumerate}

\item Additionally, if you are using existing assets (e.g., code, data, models) or curating/releasing new assets, \textbf{without compromising anonymity}...
\begin{enumerate}
  \item If your work uses existing assets, did you cite the creators?
    \answerYes{Yes, see Section 3 to 8}
  \item Did you mention the license of the assets?
    \answerNo{No}
  \item Did you include any new assets in the supplemental material or as a URL?
    \answerNA{NA}
  \item Did you discuss whether and how consent was obtained from people whose data you're using/curating?
    \answerYes{Yes, see Section 3 to 8}
  \item Did you discuss whether the data you are using/curating contains personally identifiable information or offensive content?
    \answerNo{No}
\item If you are curating or releasing new datasets, did you discuss how you intend to make your datasets FAIR?
\answerNA{NA}
\item If you are curating or releasing new datasets, did you create a Datasheet for the Dataset? 
\answerNA{NA}
\end{enumerate}

\item Additionally, if you used crowdsourcing or conducted research with human subjects, \textbf{without compromising anonymity}...
\begin{enumerate}
  \item Did you include the full text of instructions given to participants and screenshots?
    \answerNA{NA}
  \item Did you describe any potential participant risks, with mentions of Institutional Review Board (IRB) approvals?
    \answerNA{NA}
  \item Did you include the estimated hourly wage paid to participants and the total amount spent on participant compensation?
    \answerNA{NA}
   \item Did you discuss how data is stored, shared, and deidentified?
   \answerNA{NA}
\end{enumerate}

\end{enumerate}

\appendix

\section*{Appendix}

\subsection{Qualitative Studies with ChatGPT API}
\label{sec:chatgpt_api}

\subsubsection{Emoji Semantics}
We repeated the experiments in Section \ref{sec:emoji_semantic} using the ChatGPT API and found that nearly all responses were consistent with those presented in Figures \ref{fig:emoji_meaning} and \ref{fig:special_meaning}. For example, in the last example in Figure \ref{fig:special_meaning}, the ChatGPT API continues to interpret the \MyEmoji{\snake} emoji as representing untrustworthy people rather than as an animal.
Furthermore, for the emojis \MyEmoji{\slightsmilingface}, \MyEmoji{\doge}, \MyEmoji{\cowface}, and \MyEmoji{\beermug} in Section \ref{sec:semantic_cross_cultural}, \MyEmoji{\cowface} and \MyEmoji{\beermug} received correct interpretations when prompted in Chinese, while \MyEmoji{\slightsmilingface} and \MyEmoji{\doge} returned the expected responses when ChatGPT was asked to consider Chinese pop culture. This aligns with the patterns observed in Section \ref{sec:semantic_cross_cultural}.

\subsubsection{Emoji Usage Associated with Communities}
We also repeated the experiments from Section \ref{sec:emoji_community} using the ChatGPT API, and the results are presented in Tables \ref{tab:emoji_gender_chatgpt}, \ref{tab:emoji_hashtag_chatgpt}, and \ref{tab:emoji_github_chatgpt}. The model's responses from the ChatGPT API with a temperature of 0 closely align with the results obtained via the web interface in Section \ref{sec:emoji_community}. Therefore, the conclusions that ChatGPT relies on stereotypical associations to infer emoji usage patterns across different communities, as discussed in Section \ref{sec:emoji_community}, remain valid.

\begin{table}[!htb]
    \small
    \centering
    \resizebox{\columnwidth}{!}{%
    \begin{tabular}{lcc}
    \toprule
    User         & Emoji from actual usage & Emoji from ChatGPT \\ \midrule
    \begin{tabular}[c]{@{}l@{}l@{}} French user \\ \cite{lu2016emojiusgae} \end{tabular}  & \begin{tabular}[c]{@{}l@{}l@{}} \MyEmoji{\heart} \MyEmoji{\joy} \MyEmoji{\heartwitharrow} \MyEmoji{\sparklingheart} \MyEmoji{\kiss} \\ \MyEmoji{\hearteyes} \MyEmoji{\twohearts} \MyEmoji{\revolvinghearts} \MyEmoji{\purpleheart} \MyEmoji{\growingheart} \end{tabular} & \begin{tabular}[c]{@{}l@{}l@{}} \MyEmoji{\franceflag} \MyEmoji{\wineglass} \MyEmoji{\baguette} \MyEmoji{\cheese} \MyEmoji{\tower} \\ \MyEmoji{\palette} \MyEmoji{\croissant} \MyEmoji{\clinking}  \MyEmoji{\cyclist} \MyEmoji{\rooster}  \end{tabular}                        \\ \midrule
    \begin{tabular}[c]{@{}l@{}l@{}} Female user \\ \cite{chen2018gender} \end{tabular}     &  \begin{tabular}[c]{@{}l@{}l@{}} \MyEmoji{\womanholdinghands} \MyEmoji{\birthdaycake} \MyEmoji{\partypopper}  \MyEmoji{\gift} \MyEmoji{\kissmark} \\ \MyEmoji{\balloonemoji} \MyEmoji{\cherryblossom} \MyEmoji{\twohearts} \MyEmoji{\purpleheart} \MyEmoji{\nailpolish} \end{tabular} & \begin{tabular}[c]{@{}l@{}l@{}} \MyEmoji{\womanofficeworker}  \MyEmoji{\tippinghandwoman} \MyEmoji{\womansinger} \MyEmoji{\womanclothes} \MyEmoji{\nailpolish} \\ \MyEmoji{\highheeled} \MyEmoji{\womansteamy} \MyEmoji{\familywomangirlboy} \MyEmoji{\cherryblossom} \MyEmoji{\ribbon} \end{tabular}   \\ \midrule
    \begin{tabular}[c]{@{}l@{}l@{}} Male user \\ \cite{chen2018gender} \end{tabular}   & \MyEmoji{\socceremoji} \MyEmoji{\cigarette} \MyEmoji{\malesign} & \begin{tabular}[c]{@{}l@{}l@{}} \MyEmoji{\maninsuit} \MyEmoji{\beardman} \MyEmoji{\flexed} \MyEmoji{\beermug} \MyEmoji{\football} \\ \MyEmoji{\automobile} \MyEmoji{\videogame} \MyEmoji{\hammerwrench} \MyEmoji{\manlifting} \MyEmoji{\cutofmeat}  \end{tabular}       \\ \bottomrule
    \end{tabular}
    }
    \caption{Outputs from ChatGPT API (gpt4-1106-preview) and emojis from actual usage \cite{lu2016emojiusgae, chen2018gender} associated with French users, female users, and male users.}
    \label{tab:emoji_gender_chatgpt}
\end{table}

\begin{table}[!htb]
    \small
    \centering
    \resizebox{\columnwidth}{!}{%
    \begin{tabular}{lcc}
    \toprule
    Hashtag         & Emoji from actual usage & Emoji from ChatGPT \\ \midrule
    \#BlackOutBTS   &  \MyEmoji{\sparkles} \MyEmoji{\relieved} \MyEmoji{\pleadingface} \MyEmoji{\heart} \MyEmoji{\blackheart}                                    &    \MyEmoji{\musicalnote} \MyEmoji{\purpleheart} \MyEmoji{\selfie} \MyEmoji{\microphone} \MyEmoji{\fistraised}                            \\ \midrule
    \#MothersDay    &   \MyEmoji{\heart} \MyEmoji{\twohearts} \MyEmoji{\bouquet} \MyEmoji{\whiteheart} \MyEmoji{\dizzy}                                   &                          \MyEmoji {\familywomangirlboy} \MyEmoji{\bouquet} \MyEmoji{\breastfeeding} \MyEmoji{\heart} \MyEmoji{\gift}    \\ \midrule
    \#TheLastDance  &  \MyEmoji{\goat} \MyEmoji{\joy} \MyEmoji{\popcorn} \MyEmoji{\snake} \MyEmoji{\fire}                                    &                                \MyEmoji{\basketball} \MyEmoji{\trophy} \MyEmoji{\TV} \MyEmoji{\mandancing} \MyEmoji{\astonishedface} \\ \bottomrule
    \end{tabular}
    }
    \caption{Outputs from ChatGPT API (gpt4-1106-preview) and emojis from actual usage \cite{zhou2022emoji} associated with different hashtags.}
    \label{tab:emoji_hashtag_chatgpt}
\end{table}

\begin{table}[!htb]
    \small
    \centering
    \begin{tabular}{ l  l }
        \toprule
        Source & Emojis associated with GitHub  \\
        \midrule
        \multirow{2}{*}{Actual usage} &  \MyEmoji{\thumbsup} \MyEmoji{\stopsign} \MyEmoji{\smilingeyes} \MyEmoji{\rocket} \MyEmoji{\wavinghand} \MyEmoji{\partypopper} \MyEmoji{\whitecheckmark} \MyEmoji{\bugemoji} \MyEmoji{\warningemoji} \MyEmoji{\thinkingface}  \\
        & \MyEmoji{\sparkles} \MyEmoji{\slightsmilingface} \MyEmoji{\heart} \MyEmoji{\grinningface} \MyEmoji{\ladybugemoji} \MyEmoji{\smilingfacesmilingeyes} \MyEmoji{\moneybag} \MyEmoji{\pray} \MyEmoji{\triangular} \MyEmoji{\bigeyes}\\
        \midrule
        \multirow{2}{*}{ChatGPT} &  \MyEmoji{\thumbsup} \MyEmoji{\thumbsdown} \MyEmoji{\smilingeyes}  \MyEmoji{\confusedface} \MyEmoji{\partypopper} \MyEmoji{\sparkles} \MyEmoji{\bugemoji} \MyEmoji{\fire} \MyEmoji{\rocket} \MyEmoji{\thinkingface} 
        \\
        &
        \MyEmoji{\hundredpoints} \MyEmoji{\pray} \MyEmoji{\hammerwrench} \MyEmoji{\memoemoji} \MyEmoji{\penguin} \MyEmoji{\construction} \MyEmoji{\packageemoji} \MyEmoji{\globeemoji} \MyEmoji{\counterclockwisearrowsbutton} \MyEmoji{\chart}
        \\
        \bottomrule
    \end{tabular}
    \caption{Outputs from ChatGPT API (gpt4-1106-preview) and emojis from actual usage \cite{zhou2023emoji} associated with the GitHub platform.}
    \label{tab:emoji_github_chatgpt}
\end{table}

\subsection{Llama3 Experiments}
\label{sec:llama3}

To probe the open-source large language models' capabilities on emoji-related tasks, we choose the state-of-the-art (SoTA) models Llama3-8B \cite{touvron2023llama} to replicate the experiments in Section \ref{sec:emoji_semantic}, \ref{sec:emoji_sentiment}, \ref{sec:emoji_intention} and \ref{sec:emoji_community}. 

\subsubsection{Emoji Semantics}
We first ask Llama3 to explain the emoji semantics without context as Section \ref{sec:emoji_semantics_no_context} but observe that it usually generates hallucinations for the outputs. For example, for the cases in Figure \ref{fig:emoji_meaning}, Llama3 generates the correct explanation for \MyEmoji{\shrugging}, where the detailed output is: \textit{I think it is a shrug emoji. It is used to express that you don't know or don't care about something}. However, for \MyEmoji{\purpleheart} and \MyEmoji{\blueheart}, Llama3 directly outputs: \textit{I'm not sure what it means} and we do not get the desired answers for emoji semantics. 

As for the emoji semantics with context, Llama3 does not generate the correct explanations. For example, for the \MyEmoji{\honeybee} and \MyEmoji{\snake} emojis in Figure \ref{fig:special_meaning}, Llama3 will directly state that \textit{The emoji is used to represent the bee or snake}, rather than analyzing the meaning of the emoji according to the context.

\subsubsection{Emoji Sentiment}
For emoji sentiment, we duplicate the experiment for annotating the sentiment of the same 50 emojis in Section \ref{sec:emoji_sentiment} and find that Llama3 labels the same sentiment as human annotators in 41/50 cases, which has a lower accuracy than the ChatGPT annotation (49/50). 

\subsubsection{Emoji Intention}
We also repeated the experiments in Section \ref{sec:emoji_intention_wo_context} using the same prompt to ask the Llama3 model to annotate emoji intentions. However, we found that Llama3 only produced a single token: ``\textbackslash'' and then ended the generation. This indicates that Llama3 struggles to follow long instructions with intention rubrics, resulting in meaningless output.

\subsubsection{Emoji Usage Associated with Communities}
To probe whether Llama3 encodes the knowledge of associations between emojis and communities, we utilize the same prompt in Section \ref{sec:emoji_community} and ask Llama3 to generate 10 emojis associated with different communities. However, we find that, in addition to the hashtag \#TheLastDance, where it generates a basketball emoji \MyEmoji{\basketball}, it produces meaningless hallucinations, such as the code to calculate associations.

Overall, Llama3-8B's performance falls short compared to human annotators in nearly all emoji-related tasks. The reasons for its shortcomings can be attributed to two main factors. First, as observed in its explanations of emoji sentiment and semantics, Llama3 lacks the necessary knowledge of emojis. Second, due to its limited instruction-following capabilities, Llama3 often generates hallucinations when faced with long or complex prompts. Consequently, we conclude that Llama3-8B is not a suitable replacement for human annotators nor a reliable tool for answering users' questions about emojis.

\subsection{Prompt Details}

\begin{table}[!htb]
    \begin{subtable}{1\columnwidth}
    \centering
    \begin{tabular}{m{22em}}
        \toprule
        Could you assign a sentiment to this emoji: \textcolor{red}{emoji}? Please select the sentiment from positive, negative, or neutral. \\
        \bottomrule
    \end{tabular}
    \caption{Prompt for probing emoji sentiment without text context.}
    \label{tab:sentiment_no_text}
    \end{subtable}
    \bigskip
    \begin{subtable}{1\columnwidth}
    \centering
    \begin{tabular}{m{22em}}
        \toprule
        I will provide you 5 tweets. Please assign a sentiment to each tweet. 
        You can only select the sentiment from positive, negative, or neutral. 
        Please only output the sentiment answer without justification. \\
        Here are some examples:\\
        1. \@user \@user what do these '1/2 naked pics' have to do with anything? They're not even like that. \\
        2. I think I may be finally in with the in crowd \#mannequinchallenge \#grads2014 \@user \\
        3. \@user Wow,first Hugo Chavez and now Fidel Castro. Danny Glover, Michael Moore, Oliver Stone, and Sean Penn are running out of heroes. \\
        4. An interesting security vulnerability - albeit not for the everyday car thief \\
        5. Can't wait to try this - Google Earth VR - this stuff really is the future of exploration.... \\
        The sentiment output is: 
        1. neutral 
        2. positive 
        3. negative 
        4. neutral 
        5. positive \\
        Here is the tweet list: \\
        \textcolor{red}{\{tweet\}} \\
        \bottomrule
        
    \end{tabular}
    \caption{Prompt for applying ChatGPT to label the sentiment of tweets, and \textcolor{red}{\{tweet\}} is the tweet to label the sentiment.}
    \label{tab:sentiment_prediction}
    \end{subtable}
    \caption{Prompts of sentiment-related tasks for emojis.}
    \label{table:sentiment_prompt}
\end{table}

\begin{table}[!htb]
    \begin{tabular}{m{22em}}
        \toprule
        Please rate the willingness to use the emoji \textcolor{red}{\{emoji\}} for the following 7 intentions: expressing sentiment, strengthening expression, adjusting tone, expressing humor, expressing irony, expressing intimacy, describing content, on a 7-point scale (7 = totally willing, 1 = not willing at all).
        The details of the intentions are suggested as below: \\
        \textcolor{red}{\{definition details\}}
        \\
        \bottomrule
    \end{tabular}
    \caption{Prompt for probing the intention of emojis without given the text context. \textcolor{red}{definition details} can be found in \citet{hu2017spice}.}
    \label{table:intention_prompt_no_context}
\end{table}

\begin{table}[!htb]
    \centering
    \begin{tabular}{m{22em}}
        \toprule
        Please assign an intention of emoji usage from the following 8 intentions: sentiment expressed, 
        sentiment strengthened, tone adjusted, content described, content organized, content emphasized, non-communication use, symbol. 
        Please only output the intention answer without justification.
        
        The details of the intentions are suggested as below: \\
        \textcolor{red}{\{definition details\}} \\

        Here are some examples: \\
        1. Even the Guardian has TLS now.... \MyEmoji{\thumbsup} \\
        2. merge when we receive \MyEmoji{\thumbsup}, shipit \\
        3. Thanks for your effort \MyEmoji{\thumbsup} \\
        4. \MyEmoji{\checkmarkemoji} Creating directory \\
        5. Remote Get: \MyEmoji{\checkmarkemoji} \\
        The intention output is: 
        1. sentiment expressed 
        2. symbol 
        3. sentiment strengthened 
        4. non-communication use 
        5. content described \\
        
        Here is the Github post list: 
        \textcolor{red}{\{post\}} \\
        \bottomrule
        
    \end{tabular}
    \caption{Prompt for applying ChatGPT to label the intention of emojis in GitHub posts. \textcolor{red}{\{post\}} is the GitHub to label the emoji intention. \textcolor{red}{definition details} can be found in \citet{lu2018first}.}
    \label{table:github_intention_prompt}
\end{table}

\begin{table}[!htb]
    \centering
    \begin{tabular}{m{22em}}
        \toprule
        I will provide you 5 tweets. Please tell me whether the tweets are ironic or not.
        Please only output the irony label without justification.
        Here are some examples: \\
        1. Such .   You still have to \#praisehim ;) \\
        2. Hey heyy!!!I....wanna be a rockstar \#vscocam  hero \#vsocam \#hero \#spiderman \\
        3. \@user  nice to see the ambulance service is so important to OUR mps \\
        4. I love finals week! \#justkidding  \#stressed \\
        5. Literally cried when I woke up because I know what this day has in store for me \#TheStartOfTechWeek  Ready \#JustShootMeKnow \\
        The irony output is:
        1. ironic 
        2. not ironic 
        3. ironic 
        4. ironic 
        5. not ironic \\
        Here is the tweet list:
        \textcolor{red}{\{tweet\}} \\
        \bottomrule
        
    \end{tabular}
    \caption{Prompt for applying ChatGPT to label the irony or not of a tweet. \textcolor{red}{\{tweet\}} is the tweet to label the irony.}
    \label{table:tweet_irony}
\end{table}

\begin{table}[!htb]
    \centering
    \begin{tabular}{m{22em}}
        \toprule
        I will provide you 5 tweets with emojis in Arabic language. Please tell me which type of irony the emojis in tweets convey.
        You can only select the irony type from humour, sarcasm, and no irony.
        Here is the definition of sarcasm and humour:
        Sarcasm, a specific facet of irony, employs language with a sharp, often bitter tone to mock or critique, often with exaggerated emphasis.
        Humour, while encompassing elements of irony, specifically denotes the ability to evoke amusement or laughter.
        Please only output the irony label without justification.
        Here are some examples: \\
        1. \textcolor{red}{\{Arabic tweet example 1\}} \\
        2. \textcolor{red}{\{Arabic tweet example 2\}} \\
        3. \textcolor{red}{\{Arabic tweet example 3\}} \\
        4. \textcolor{red}{\{Arabic tweet example 4\}} \\
        5. \textcolor{red}{\{Arabic tweet example 5\}} \\
        The irony output is:
        1. no irony
        2. no irony
        3. sarcasm
        4. sarcasm
        5. humour \\
        Here is the tweet list:
        \textcolor{red}{\{tweet\}} \\
        \bottomrule
        
    \end{tabular}
    \caption{Prompt for applying ChatGPT to label the irony of emojis in Arabic tweets. \textcolor{red}{\{Arabic tweet example n\}} is the $n$-th example of an Arabic tweet that contains emojis with the ground-truth irony label for the emoji. \textcolor{red}{\{tweet\}} is the tweet that contains emojis to label the irony.}
    \label{table:irony_arabic}
\end{table}

\begin{table}[!htb]
    \centering
    \begin{tabular}{m{22em}}
        \toprule
        I will provide you 5 tweets. Please recommend only one emoji for each tweet.
        You can only select emojis from these emoji candidates: \textcolor{red}{\{emoji candidates\}}.
        Please only output the emoji answer without justification.
        Here are some examples: \\
        1. Sunday afternoon walking through Venice in the sun with @user \\
        2. Time for some BBQ and whiskey libations. Chomp, belch, chomp! (@user) \\
        3. Man these are the funniest kids ever!! That face! \#HappyBirthdayBubb @user Xtreme \\
        4. \#sandiego @San Diego, California \\
        5. My little \#ObsessedWithMyDog @user Capitol Hill \\
        The emoji recommendation is: 
        1. \MyEmoji{\sunemoji} 
        2. \MyEmoji{\smirk} 
        3. \MyEmoji{\smilingfacewithtear} 
        4. \MyEmoji{\usaflag} 
        5. \MyEmoji{\heart} \\
        Here is the tweet list: 
        \textcolor{red}{\{tweet\}} \\
        \bottomrule
        
    \end{tabular}
    \caption{Prompt for applying ChatGPT to predict the emoji in tweets. \textcolor{red}{\{tweet\}} is the tweet to predict the emojis. \textcolor{red}{\{emoji candidates\}} represents the candidate emojis that ChatGPT should select from and the details of candidate emojis can be found in \citet{barbieri2018semeval}.}
    \label{table:emoji_prediction}
\end{table}


\end{document}